\documentclass[10pt,twocolumn,letterpaper]{article}

\usepackage{cvpr}              

\usepackage{graphicx}
\usepackage{amsmath}
\usepackage{amssymb}
\usepackage{booktabs}
\usepackage{enumitem}

\usepackage[pagebackref,breaklinks,colorlinks]{hyperref}

\usepackage[capitalize]{cleveref}
\crefname{section}{Sec.}{Secs.}
\Crefname{section}{Section}{Sections}
\Crefname{table}{Table}{Tables}
\crefname{table}{Tab.}{Tabs.}

\def\cvprPaperID{7408} 
\def\confName{CVPR}
\def\confYear{2023}

\graphicspath{{figs/}}

\begin{document}

\title{ FEND: A Future Enhanced Distribution-Aware Contrastive Learning Framework for Long-tail Trajectory Prediction}

\author{~Yuning Wang{\small $~^{1}$}\footnotemark[1], ~Pu Zhang{\small $~^{2}$}\footnotemark[1], ~Lei Bai {\small $~^{3}$}, ~Jianru Xue{\small $~^{1}$}\footnotemark[2]\\
\normalsize
$^{1}$\
Institute of Artificial Intelligence and Robotics, Xi'an Jiaotong University, China\\
\normalsize
$^{2}$\,  DiDi Chuxing, China\\
\normalsize
$^{3}$\, Shanghai AI Laboratory, China\\
\normalsize
\normalsize
{wangyn}@stu.xjtu.edu.cn,
\normalsize
\{zhangpu94,baisanshi\}@gmail.com,
\normalsize
jrxue@mails.xjtu.edu.cn
}

\maketitle

\renewcommand{\thefootnote}{\fnsymbol{footnote}}
\footnotetext[1]{Equal contributions.}
\footnotetext[2]{Corresponding author.}
\begin{abstract}
Predicting the future trajectories of the traffic agents is a gordian technique in autonomous driving. However, trajectory prediction suffers from data imbalance in the prevalent datasets, and the tailed data is often more complicated and safety-critical. In this paper, we focus on dealing with the long-tail phenomenon in trajectory prediction. Previous methods dealing with long-tail data did not take into account the variety of motion patterns in the tailed data. In this paper, we put forward a future enhanced contrastive learning framework to recognize tail trajectory patterns and form a feature space with separate pattern clusters.Furthermore, a distribution aware hyper predictor is brought up to better utilize the shaped feature space.
Our method is a model-agnostic framework and can be plugged
 into many well-known baselines. Experimental results show that our framework outperforms the state-of-the-art long-tail prediction method on tailed samples by 9.5\% on ADE and 8.5\% on FDE, while maintaining or slightly improving the averaged performance. Our method also surpasses many long-tail techniques on trajectory prediction task.
\end{abstract}


\section{Introduction}
\label{sec:intro}

Trajectory prediction is of great importance in autonomous driving scenarios \cite{luo2018porca}. It aims to predict a series of future positions for the agents on the road given the observed past tracks. There have been many recent methods in trajectory prediction, both unimodal\cite{alahi2016social,zhang2019sr} and multimodal\cite{gupta2018social,sadeghian2019sophie,salzmann2020trajectron++,zhang2020social}.

\begin{figure}
    \centering
    \includegraphics[width=1.0\linewidth]{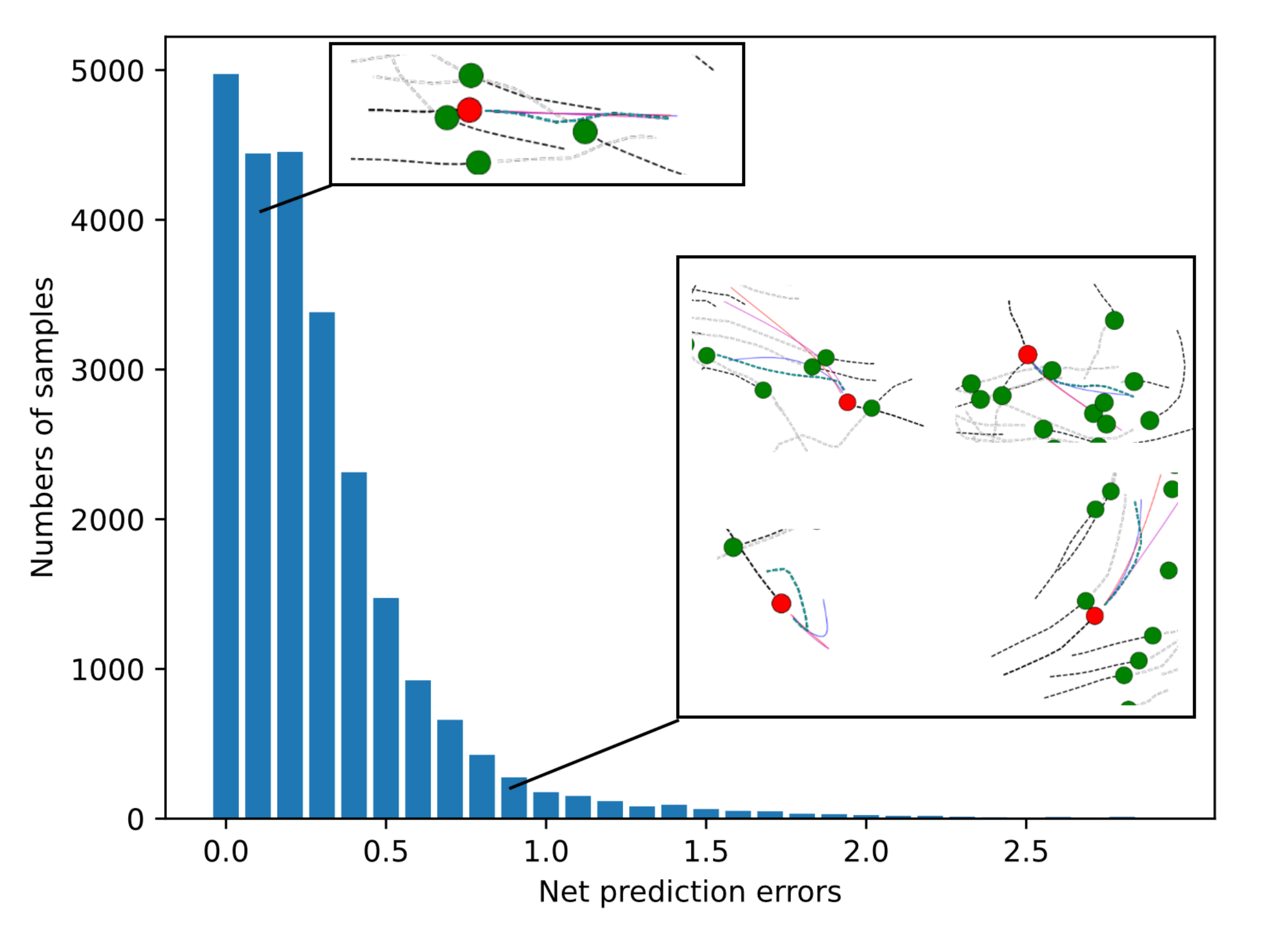}
    \caption{The long-tailed final displacement errors of the state-of-the-art prediction network: Trajectron++ EWTA \cite{makansi2021exposing} on ETH-UCY. The long-tail part of the dataset contains various complicated motion patterns, and predicting them is challenging.}
    \label{fig:test_loss}
\end{figure}

Despite the high accuracy those prediction methods have achieved, most of them treat the samples in the datasets equally in both training and evaluation phases. But there is a long-tailed phenomenon in prevalent datasets\cite{makansi2021exposing}. For example, in real traffic scenes, most of the trajectories follow certain simple kinematic rules, while deviating and collision-avoiding circumstances are scarce. Therefore, the frequent cases are often simple and easy to predict, while the tail cases are often complicated with many motion patterns and suffer from large prediction errors, which makes them more safety-critical, as shown in \cref{fig:test_loss} for the univ dataset. Despite of its significance, the long-tail prediction problem have been rarely discussed in literature.

It has been pointed out that the feature encoders largely suffer from long-tail data. In the training process, the head samples are encountered more often and dominate the latent space, while the tailed samples will be modeled insufficiently, as discussed in \cite{samuel2021distributional,li2022targeted,makansi2021exposing}. Feature embeddings of the tailed data can even be mixed up with the ones of the head data as 
discussed in \cite{makansi2021exposing}, therefore the performances of the tailed samples could be harmed.

In this paper, we pick up the general idea of using contrastive learning to enhance the model ability on long-tailed data. A new framework is developed called FEND: Future ENhanced Distribution-aware contrastive trajectory prediction, which is a pattern-based contrastive feature learning framework enhanced by future trajectory information. An offline trajectory clustering process and prototypical contrastive learning are introduced for recognizing and separating different trajectory patterns to boost the tail samples modeling. To deal with the afore mentioned problem, the features of trajectories within the same pattern cluster are pulled together, while the features from different pattern clusters will be pushed apart. Moreover, a more flexible network structure of the decoder is introduced to exploit the shaped feature embedding space with different pattern clusters. 
Our contribution can be summarized as follows:

\begin{itemize}[noitemsep,topsep=0pt,parsep=0pt,partopsep=0pt]
\item 
We propose a future enhanced contrastive feature learning framework for long-tailed trajectory prediction, which can better distinguish tail patterns from head patterns, and the different patterns are represented by different cluster prototypes to enhance the modeling of the tailed data.

\item 
We propose a distribution-aware hyper predictor, aiming at providing separated decoder parameters for trajectory inputs with different patterns.

\item 
Experimental results show that our proposed framework can outperform start-of-the-art methods.

\end{itemize}

\footnotetext{Codes available at https://github.com/ynw2021/FEND.}

\section{Related Work}
\label{sec:related}

\subsection{Trajectory Prediction}
Deep learning has become a mainstream trajectory prediction method because of its powerful representational ability. Some studies \cite{alahi2016social,zhang2019sr,mohamed2020social,xu2022groupnet,yu2020spatio} focus on better modeling subtle relationship such as social interactions to make their prediction more precise, and some works \cite{narayanan2021divide,makansi2019overcoming,pang2021trajectory,phan2020covernet,zhao2021tnt} aim to produce more diverse trajectory proposals. Strong baselines \cite{salzmann2020trajectron++,mangalam2021goals,yuan2021agentformer,shafiee2021introvert} have been brought up. Although the trajectory prediction methods become increasingly accurate, the long-tail issue in the task of trajectory prediction has been rarely discussed.

\textbf{Trajectory prediction approaches based on clustering}.
Existing methods \cite{sun2021three,xu2022remember,chai2019multipath} have used trajectory clustering for trajectory prediction. MultiPath \cite{chai2019multipath} performs Kmeans with the square distances between the trajectories to get anchor trajectory sets. PCCS-Net\cite{sun2021three} decouples multimodal trajectory prediction into three steps: feature clustering, cluster selecting, and synthesizing. Memo-Net\cite{xu2022remember} clusters trajectories in the original coordinates and uses an attention network for better cluster selecting. All existing methods that use trajectory clustering are 
aiming at selecting future modalities for trajectory decoders and producing more diverse trajectories, which is different from our goal to distinguish tail patterns from head patterns and optimize the feature embedding space.

\textbf{Trajectory prediction approaches based on contrastive learning}.
Contrastive learning \cite{oord2018representation} is a self-supervised method to improve the representation ability of the network given the similarities between sample pairs, and has many variants \cite{chen2020simple,caron2020unsupervised,khosla2020supervised,kalantidis2020hard} with different ways of selecting positive and negative samples and calculating contrastive loss. Prototypical Contrastive Learning (PCL) \cite{li2020prototypical} is a variant of contrastive learning that can preserve local smoothness therefore induce semantically hierarchical clustered feature space\cite{li2020prototypical}. 
Contrastive learning has also been incorporated into trajectory prediction. DisDis\cite{chen2021personalized} uses contrastive learning in a CVAE framework to discriminate the latent variable distributions and make the predictions more diverse. ABC+\cite{halawa2022action} uses action labels from their datasets and contrasts according to them. Social-NCE\cite{liu2021social} uses contrastive learning to make the predictions away from their simulated collision cases. None of those above-mentioned methods have discussed long-tail prediction. The most relevant work is from Makansi \textit{et al.} \cite{makansi2021exposing}, which also tries to solve the long-tail prediction problem with contrastive learning and uses Kalman prediction errors to select positive and negative samples. Makansi \textit{et al.} \cite{makansi2021exposing} push all the tailed samples together in their method.
In this work, we not only separate the tails from the heads as the study\cite{makansi2021exposing} did, but also recognize the patterns of the tailed samples due to the fact that the tailed samples can be tailed in different ways, \eg. turning or accelerating, as shown in \cref{fig:test_loss} and \cref{fig:qualitative}, which further improves the model capabilities.

\subsection{Long-tailed Learning}
Long-tailed learning aims to improve the performance on tailed samples when faced with unbalanced data. Most of them focus on classification tasks. Typical methods do data resampling\cite{shen2016relay,chawla2002smote,han2005borderline} or loss reweighting\cite{cui2019class,he2009learning,lin2017focal} to improve the capability of the network on tailed samples. Recent advances\cite{cao2019learning,menon2020long} seek for a theoretical balance of head-tail performance by means of adjusting the classification boundaries, whereas these methods cannot be directly used in regression tasks. Very recently Yang \textit{et al.} \cite{yang2021delving} have investigated imbalanced regression tasks and propose a feature distribution smoothing and label distribution smoothing method. But the methodology in \cite{yang2021delving} needs labels with structured relationships, which is incongruent with the trajectory data.
In our methods, we find out structured relationships between trajectories by forming pattern clusters, and optimize feature space according to item. 
Besides, we use Hypernetwork \cite{ha2016hypernetworks} as the trajectory decoder to deal with tail samples utilizing its distribution-aware modeling ability, which has not been discussed in long-tail regression to our best knowledge.

\begin{figure*}[h]
\small
    \centering
    \includegraphics[width=0.85\linewidth]{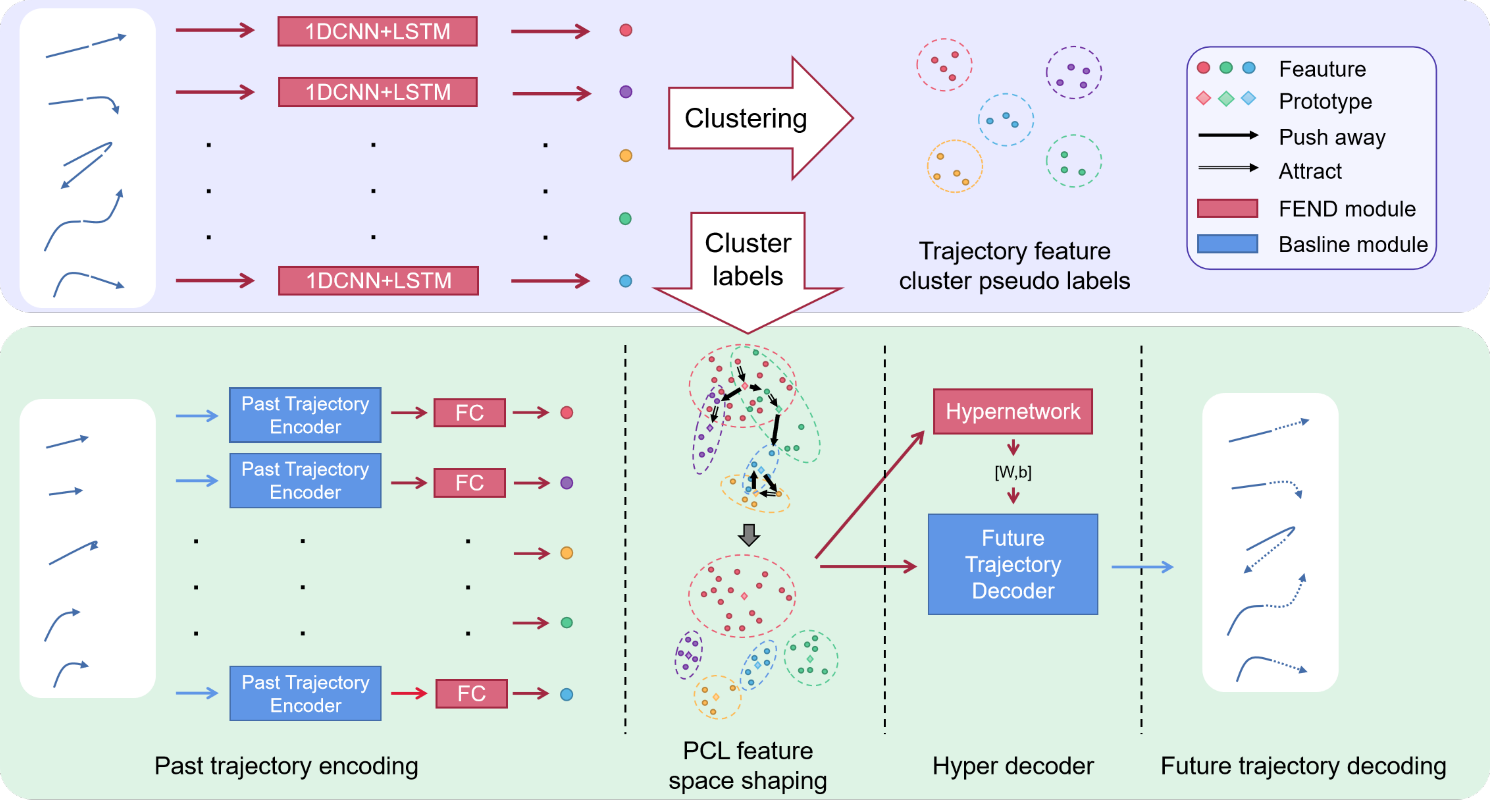}
    \caption{Illustration of our overall future enhanced distribution-aware contrastive learning framework. Top: Offline Kmeans clustering for pseudo cluster labels. Bottom: The baseline prediction network with FEND plugged in for prediction. The FEND module contains a PCL optimization procedure and a hyper decoder. }
    \label{fig:framework}
\end{figure*}

\section{Method}
\label{sec:methods}

\textbf{Problem formulation}. Trajectory prediction is a kind of sequential prediction problems. Given a series of past observed coordinates $\{(x^n_t,y^n_t)\}^N_{n=1}$ for $N$ agents over time $ t=-T_{obs}+1 ,-T_{obs}+2,...,0$, our objective is to predict the future locations $(\hat{x}_t,\hat{y}_t)$ of the agent of interest in a constant period $t=1,2,...,T_{pred}$.

As discussed above, the trajectory data suffers from the long-tail phenomenon. To address this issue, we come up with a new long-tail trajectory prediction framework FEND, which contains \textit{a future enhanced contrastive learning method} for helping shape better feature embedding for trajectory encoders, and \textit{a more flexible distribution-aware hyper predictor} for impairing the influence from the head samples to the tail samples.

\textbf{Overview.} The overall framework of FEND is discribed in \cref{fig:framework}. Both the history and the future trajectories are firstly processed by a trajectory feature extractor, and the extracted features are clustered by Kmeans to form different pattern clusters. After clustering, the tail trajectory patterns and the head trajectory patterns are separated spontaneously using both history and future information. 
According to the pseudo cluster labels generated by Kmeans, PCL is performed on the history encoding features of the baseline prediction network. By performing PCL, the feature space of trajectory encoders is separately clustered. 
Then a hyper decoder is constructed which generates separate decoder weights for different trajectory inputs, therefore trajectories in the head clusters and the tail clusters are predicted differently.

\subsection{Future Enhanced Contrastive Learning}

  \subsubsection{Future Enhanced Trajectory Clustering}
  \label{sec:clustering}
  
    For trajectory pattern clustering, the start points and initial directions of trajectories should be normalized to make the feature extractor more focused on the different patterns of trajectories. But the data-preprocessing ways of present trajectory prediction methods are various. Therefore, to make our framework can be more generally applied, we use an offline cluster module to do normalization and perform trajectory clustering. Also, many trajectory prediction baselines do not have a future trajectory encoder, and their encoded past trajectory features are high-dimensional, so online clustering will be time-consuming.
    
    We simply use a 1D convolution network (CNN) attached by an LSTM as the trajectory feature extractor for trajectory encoding and reconstruction, which is supervised by the reconstruction loss as autoencoders. The feature at the bottleneck of the network is used to perform hierarchical Kmeans. Kmeans \cite{hartigan1979algorithm} is a computation-efficient classical clustering method and can be replaced with any other clustering algorithm. We perform Kmeans with multiple level of clusters for achieving hierarchy, as the original PCL does. In the training phase of feature extractors, we also use the original PCL \cite{li2020prototypical} with EM steps as an auxiliary loss to get a hierarchically clustered feature space in a self-supervised manner, which will be discussed in \cref{sec:PCL}.
    
   \subsubsection{Prototypical Contrastive Learning}
   \label{sec:PCL}

    In our methods, we have already got the cluster labels after the trajectory clustering step. Therefore we use the cluster assignments as pseudo labels for computing prototypes and densities. The original PCL \cite{li2020prototypical} is an self-supervised methods with EM steps, therefore it needs to perform clustering before every training epoch. Our methods use the pseudo labels to reduce the clustering steps therefore require less computation source compared to the original PCL. Given pseudo cluster labels, PCL can pull the features of instances belonging to the same cluster together and push the features of instances in different clusters apart, as what vanilla contrastive learning does to the positive and negative samples. 
    
    \textbf{Implementing PCL loss. }
    We do PCL on the features at the bottleneck of the encoder-decoder trajectory prediction network: after the encoder. Similar to Makansi \textit{et al.} \cite{makansi2021exposing}, we add a fully-connected (FC) layer after the encoder and add the PCL loss to its output features. The features before the FC layer will be given to the trajectory decoder. We perform a multi-level clustering with $M$ hierarchies when calculating PCL loss. The PCL loss is as follows:
    
\begin{equation}
\begin{aligned}
\mathcal{L}_{ProtoNCE}=\mathcal{L}_{ins}+\mathcal{L}_{proto},
\end{aligned}
\label{eq:PCL}
\end{equation}
where the first term is an instance-wise contrastive term and the second-term is an instance-prototype contrastive term.

    \textbf{Instance-wise term.}
    The first term  in \cref{eq:PCL} is an instance-wise contrastive term considering the pseudo cluster labels, which can be written as follows:
    \begin{equation}
    \begin{aligned}
    \mathcal{L}_{ins} =-\sum_{i=1}^r\frac{1}{N_{\mathbf{p o}_i}} \sum_{i_+=1}^{N_{\mathbf{po}_i}}\log \frac{\exp \left(v_i \cdot v_{i_+} / \tau\right)}{\sum_{j=1}^r \exp \left(v_i \cdot v_j / \tau\right)}.
    \end{aligned}
    \label{eq:instance}
    \end{equation}
    The instance-wise term can help the instances gather together faster and the algorithm converge faster. $v_i$ and $v_{i_+}$ are feature embeddings of trajectory instance $i$ and positive sample $i+$ after the encoder respectively,  $i_+\neq i$. $N_{\mathbf{p o}_i}$ is the number of positive samples to $i$ in a batch. $\tau$ is the contrastive temperature of the instance-wise contrastive term. In \cref{eq:instance}, the positive samples $i+$ are the instances from the same cluster with the instance $i$, and the rest instances in the batch, \ie belonging to other clusters, are regarded as negative samples. $j$ means an arbitrary sample in the current batch data. $r$ denotes the batch size.
    
    \textbf{Instance-prototype term.}
    The second term in \cref{eq:PCL} is an instance-prototype contrastive term, which can be written as follows:
    \begin{equation}
    \begin{aligned}
    \mathcal{L}_{proto} =-\frac{1}{M} \sum_{i=1}^r \sum_{m=1}^M \log \frac{\exp \left(v_i \cdot c_s^m / \phi_s^m\right)}{\sum_{j=1}^{N_m} \exp \left(v_i \cdot c_j^m / \phi_j^m\right)}.
    \end{aligned}
    \label{eq:proto}
    \end{equation}
    
    The prototypes help preserving local smoothness and the formation of clusters with different patterns. In \cref{eq:proto}, $M$ is the number of Kmeans clustering hierarchies, $c_s^m$ means the prototype of the cluster to which $i$ belongs, and $c_j^m$ means the prototype of an arbitrary cluster $j$. The prototype is calculated by taking an average of all the features in a cluster. $N_m$ denotes the number of clusters for hierarchy $m$. $\phi_j^m$ denotes the density of a cluster $j$, which is calculated as below:
    \begin{equation} 
    \phi=\frac{\sum_{z=1}^Z\left\|v_z^{\prime}-c\right\|_2}{Z \log (Z+\alpha)},
    \label{eq:density}
    \end{equation}
    where $Z$ is the number of instances in the cluster, and $\alpha$ is a smoothing factor to ensure that small clusters do not have an overly large $\phi$. We set $\alpha=10$ same as \cite{li2020prototypical}. $v_z^{\prime}$ is the momentum updated feature for instance $z$ to ensure stability.

\subsection{Distribution-Aware Hyper Predictor}
\label{subsec:hypernetwork}

\textbf{Distribution-aware hypernetwork.} Intuitively, the head clusters and the tail clusters should be assigned different decoders to impair their influence on each other. But there is an insufficient amount of data for the tail samples, and separately training decoders for them will cause badly overfitting. Therefore, we want to transfer common knowledge across the whole dataset, while keep the modeling flexibility of separate decoders. HyperNetworks\cite{ha2016hypernetworks} is an approach of using a small network, which is known as a hypernetwork, to generate the weights of the main network, and it naturally suits our demands. The hypernetwork contains the knowledge of all samples,
which prevents overfitting. Also, there are separate decoder parameters for head and tail clusters, which make the decoder \textit{aware of the distribution of the clustered feature space}. So the hyper decoder can predict the tailed clusters differently.

 \textbf{LSTM trajectory decoder.} As an example of a hyper predictor, we employ an LSTM as the trajectory decoder, which is commonly used in recent studies \cite{makansi2021exposing,salzmann2020trajectron++,zhang2020social}.
 The original formulation of an LSTM is as follows:
\begin{equation}
\begin{aligned}
i_t &=W_h^i h_{t-1}+W_x^i x_t+b^i,\\
g_t &=W_h^g h_{t-1}+W_x^g x_t+b^g, \\
f_t &=W_h^f h_{t-1}+W_x^f x_t+b^f, \\
o_t &=W_h^o h_{t-1}+W_x^o x_t+b^o, \\
m_t &=\sigma\left(f_t\right) \odot m_{t-1}+\sigma\left(i_t\right) \odot \psi \left(g_t\right), \\
h_t &=\sigma\left(o_t\right) \odot \psi\left(m_t\right),
\end{aligned}
\label{eq:LSTM}
\end{equation}
where $i,g,f,o$ are the input gate, update gate, forget gate, and output gate respectively. $W_h \in \mathbb{R}^{N_h\times N_h},W_x \in \mathbb{R}^{N_h\times N_x},b \in \mathbb{R}^{N_h}$, $N_h$ and $N_x$ are the dimensions of input and hidden states. $h_t,m_t$ are the hidden state and the cell state. $\sigma$ is the \textit{sigmoid} operator, and $\psi$ is the \textit{tanh} operator. The initial $x$ and $h$ are produced by the feature embedding $v$ of the observed trajectory:
\begin{equation}
\begin{aligned}
x_1=W^v_xv+b^v_x,\\
h_0=W^v_hv+b^v_h,
\end{aligned}
\end{equation}
where $W^v_h \in \mathbb{R}^{N_h\times N_v},W^v_x \in \mathbb{R}^{N_x\times N_v}, b^v_h \in \mathbb{R}^{N_h}, b^v_x \in \mathbb{R}^{N_x}$. 

\textbf{HyperLSTM.} In our implement, the formulation of an LSTM with a small hypernetwork is as follows:
\begin{equation}
\begin{aligned}
y_t &=L N\left(d_h^y \odot W_h^y h_{t-1}+d_x^y \odot W_x^y x_t+b^y\left(z_b^y\right)\right),
\end{aligned}
\label{eq:hyperlstm}
\end{equation}
where
\begin{equation}
\begin{aligned}
d_h^y\left(z_h\right) &= W_{h z}^y z_h, \\
d_x^y\left(z_x\right) &= W_{x z}^y z_x, \\
b^y\left(z_b^y\right) &= W_{b z}^y z_b^y+b_0^y.
\end{aligned}
\label{eq:hyperweight}
\end{equation}

In \cref{eq:hyperlstm}, $y$ means one of $\{i,g,f,o\}$ four gates in the original LSTM formulation \cref{eq:LSTM} for brevity. $\odot$ denotes the element-wise product operation, $LN()$ denotes the layer normalization, $d$s and $b$ are the weights and bias adjusting vectors from the hypernetwork to change the weights and bias in the original LSTM. $d$s and $b$ are generated by the output $z$s of the hypernetwork as in \cref{eq:hyperweight}, where $W$s and $b_0^y$ are the weights and bias of the linear fully-connected layers. $z$ can be written as follows for instance $i$ with input feature $v_i$:

\begin{equation}
\begin{aligned}
z_i=f_{H}\left(v_i \right), \\
\end{aligned}
\label{eq:hyperoutput}
\end{equation}
where the $f_H$ means the hypernetwork mapping function, which should be a shallow network to reduce computation and prevent overfitting.

\subsection{Loss Reweighting}
\label{sec: loss reweighting}
Our final network loss is as follows:
\begin{equation}
\begin{aligned}
\mathcal{L}=\mathcal{L}_{pred} + \lambda\mathcal{L}_{ProtoNCE},
\end{aligned}
\end{equation}
where $\mathcal{L}_{pred}$ is the loss of the baseline prediction network, $\lambda$ is a coefficient on the PCL loss term. For easy samples that the network has already fitted perfectly, the PCL loss would hardly bring more benefit in network optimization. Thus, we make $\lambda$ vary across samples, which performs as a gate to shut off the PCL loss on easy samples. We use the prediction loss $\mathcal{L}_{pred}$ of the network after a warm-up training stage to indicate the hardness of the samples, which is denoted as $\mathcal{L}^{\prime}_{pred}$. $\lambda$ is determined according to $\mathcal{L}^{\prime}_{pred}$:

\begin{equation}
\begin{aligned}
\lambda&=a \quad  \quad \quad \mathcal{L}^{\prime}_{pred}>\theta,\\
\lambda&=0 \quad  \quad \quad \mathcal{L}^{\prime}_{pred}<\theta,
\end{aligned}
\end{equation}
where $a$ is a constant hyperparameter, and $\theta$ is the threshold to filter out head samples.

\section{Experiments}
\label{sec:experiments}

\begin{table*}[]
\small
\centering
\begin{tabular}{llllllll}
\toprule
& Top 1\%   & Top 2\%   & Top 3\%   & Top 4\%   & Top 5\%   & Rest        & All         \\ 
                                       
\midrule
Traj++ EWTA\cite{makansi2021exposing}                      & 0.98/2.54 & 0.79/2.07 & 0.71/1.81 & 0.65/1.63 & 0.60/1.50 & \textbf{0.14/0.26} & 0.17/0.32 \\ 
Traj++ EWTA+resample \cite{shen2016relay}            & 0.90/2.17 & 0.77/1.90 & 0.73/1.78 & 0.66/1.60 & 0.64/1.52 & 0.20/0.41 & 0.23/0.47  \\ 
Traj++ EWTA+reweighting \cite{cui2019class}        & 0.97/2.47 & 0.78/2.03 & 0.68/1.73 & 0.62/1.55 & 0.56/1.40 & 0.14/0.26 & \textbf{0.16/0.32}  \\ 
Traj++ EWTA+LDAM \cite{cao2019learning}                & 0.92/2.35 & 0.76/1.96 & 0.68/1.71 & 0.62/1.53 & 0.57/1.37 & 0.15/0.27 & 0.17/0.32 \\ 
Traj++ EWTA+contrastive\cite{makansi2021exposing}          & 0.92/2.33 & 0.74/1.91 & 0.67/1.71 & 0.60/1.48 & 0.55/1.32 & 0.15/0.27 & 0.17/0.32 \\ 
Traj++ EWTA+FEND (ours)                 & \textbf{0.84/2.13} & \textbf{0.68/1.68} & \textbf{0.61/1.46} & \textbf{0.56/1.30} & \textbf{0.52/1.19} & 0.15/0.27 & 0.17/0.32 \\ 
\bottomrule
\end{tabular}
\caption{Prediction errors in the format of (minADE/minFDE) in meters on seven kinds of testing samples on the ETH-UCY dataset.}
\label{tab:result}
\end{table*}

\begin{table*}[]
\small
\centering
\begin{tabular}{llllllll}
\toprule
                              & Top 1\%              & Top 2\%             & Top 3 \%            & Top 4\%             & Top 5\%              & Rest               & All                \\
\midrule
Traj++ EWTA \cite{makansi2021exposing}                 & 1.33/3.09          & 1.02/2.35          & 0.87/2.00          & 0.80/1.80          & 0.74/1.64          & 0.16/0.26          & 0.19/0.32          \\
Traj++ EWTA+contrastive \cite{makansi2021exposing}       & 1.28/2.85          & 0.97/2.15          & 0.83/1.83          & 0.76/1.64          & 0.70/1.48          & 0.15/0.24          & 0.18/0.30          \\
Traj++ EWTA w/o resampling+FEND & \textbf{1.21/2.50} & \textbf{0.92/1.88} & \textbf{0.79/1.61} & \textbf{0.72/1.43} & \textbf{0.66/1.31} & \textbf{0.14/0.20} & \textbf{0.17/0.26}   \\
\bottomrule
\end{tabular}
\caption{Prediction errors in the format of (minADE/minFDE) in meters on seven kinds of testing samples on Nuscenes dataset.}
\label{tab:nuscenes result}
\end{table*}

\begin{table*}[]
\small
\centering
\begin{tabular}{llllllll}
\toprule
           & Top 1\%      & Top 2\%      & Top 3\%     & Top 4\%     & Top 5\%    &Rest   & All        \\
\midrule
Y-Net* \cite{mangalam2021goals}     & 65.82/134.01 & 51.84/104.37 & 43.74/88.21 & 38.68/76.08 & 34.72/67.46 &  \textbf{6.54/8.96}  & 7.93/11.88 \\
Y-Net*+FEND & \textbf{57.58/108.51} & \textbf{46.33/86.93} & \textbf{39.22/75.02} & \textbf{35.05/66.24} & \textbf{31.27/57.98} & 6.64/9.24 & \textbf{7.87/11.68} \\
\bottomrule
\end{tabular}
\caption{Prediction errors in the format of (minADE/minFDE) on seven kinds of testing samples on SDD dataset. * means the results are reproduced using the official released code of \cite{mangalam2021goals}.}
\label{tab:sdd result}
\end{table*}

\begin{table*}
\small
\centering
\begin{tabular}{llllllllll}
\toprule
\multicolumn{3}{c}{Components} & \multicolumn{7}{c}{Performance(ADE/FDE)}                                                                                                             \\
\midrule  
 PCL                      & F  & H  & Top 1\%            & Top 2\%            & Top 3\%            & Top 4\%            & Top 5\%            & Rest                 & All                  \\ 
\midrule
     &    &    & 0.98/2.54          & 0.79/2.07          & 0.71/1.81          & 0.65/1.63          & 0.60/1.50          & \textbf{0.14/0.26} & \textbf{0.17/0.32} \\ 
\checkmark        &    &    & 0.96/2.41          & 0.79/2.03          & 0.70/1.77          & 0.62/1.56          & 0.57/1.41          & 0.15/0.27          & 0.17/0.32           \\ 
\checkmark      &\checkmark    &    & 0.89/2.23          & 0.72/1.84          & 0.66/1.61          & 0.60/1.44          & 0.55/1.30          & 0.15/0.27          & 0.17/0.32          \\ 
\checkmark      &    &\checkmark    & 0.90/2.28          & 0.72/1.87          & 0.65/1.61          & 0.58/1.43          & 0.54/1.30          & 0.15/0.27          & 0.17/0.32          \\ 
\checkmark   &\checkmark    &\checkmark    & \textbf{0.84/2.13} & \textbf{0.68/1.68} & \textbf{0.61/1.46} & \textbf{0.56/1.30} & \textbf{0.52/1.19} & 0.15/0.27          & 0.17/0.32          \\ 
\bottomrule
\end{tabular}
\caption{Ablation study of different modules in FEND. F means future enhanced clusters, H means the hypernetwork. }
\label{tab:ablation}
\end{table*}

\subsection{Datasets}

We evaluate our proposed method on several widely used public pedestrain datasets including ETH-UCY, Nuscenes and SDD. 
ETH-UCY is a pedestrian dataset with rich social interactions.
Nuscenes is a large scale trajectory dataset with both vehicles and pedestrians. 
In this work, we mainly evaluate the performances of our model on the vehicle type, same as \cite{makansi2021exposing}. 
SDD is another large scale bird view trajectory dataset. We use ETH-UCY and Nuscenes in the way same as our backbone Traj++ EWTA \cite{makansi2021exposing} and SDD in the way same as our backbone Y-Net \cite{mangalam2021goals}.

\subsection{Evaluation Metrics}

\textbf{Performance metrics.}
We use the common metrics for evaluating multimodal trajectory prediction performance: Average-Displacement-Error (ADE) and Final-Displacement-Error (FDE), which is commonly used in studies \cite{alahi2016social,zhang2019sr,chai2019multipath}. ADE means the averaged L2 distance between future prediction and ground truth trajectory, while FDE means the L2 distance between the final predicted destination and the ground truth destination. For evaluating multi-modality, we calculate mininum ADE and FDE among all the output guesses, which are denoted as minADE and minFDE and are averaged across the dataset.

\textbf{Tailed test sample selecting.}
In order to demonstrate our model on the long-tailed data, we need to separate the hard samples as well as the easy ones for evaluation. Specifically, we use the testing FDEs of the baseline method as the threshold to divide the datasets into seven kinds of samples: the top 1\%-5\% challenging samples with the largest errors, the rest easier samples, as well as all samples in the datasets.
In \cite{makansi2021exposing}, the Kalman predictor prediction error is utilized for dataset division. Compared with the FDEs of a simple Kalman predictor, performances of an advanced baseline predictor can better reflect the degrees of difficulty for the samples to be modeled by the data-driven network, which can better reveal the ability of the long-tail prediction methods to deal with the hard tailed samples. 
The Kalman divisions are discussed in supplementaries.

\subsection{Baseline}
We use Trajectron++ EWTA (Traj++ EWTA) \cite{makansi2021exposing} as a baseline for our framework on ETH-UCY and Nuscenes, which has achieved state-of-the-art results according to \cite{makansi2021exposing}. Traj++ EWTA augments the Trajectron++ \cite{salzmann2020trajectron++} by removing the conditional variational autoencoder parts and using a multi-head decoder trained with the evolving winner-take-all (EWTA) strategy.
Another strong baseline we experiment on is Y-Net\cite{mangalam2021goals}, which uses a U-Net backbone and achieves state-of-the-art results on SDD.

\subsection{Implement Details}We follow the train schedule of Traj++ EWTA, to train the network with a batch size of $256$ for $100$ epochs for ETH-UCY and 5 epochs for Nuscenes in each EWTA stage. The learning rate is initially set as $0.01$ and exponentially decays with the rate of 0.001. 
We use a warm-up of $300$ epochs in our final model for ETH-UCY. We set  $a=50$ as an initial loss factor same as \cite{makansi2021exposing}, and $a$ will decade to $0.2$ after $100$ epochs to not to harm the prediction training process, according to the drop on the EWTA loss. The head sample filter threshold $\theta$ is set to $0.2$. For the feature extractor, we use a 1D CNN with $16$ output channel and a kernel size of $3$, attached with an LSTM with a hidden size of $128$. For Kmeans clustering, we use $\{20,50,100\}$ as the cluster numbers for getting hierarchical clusters. And we use a fully-connected multilayer perception with a hidden size of $128$ as the hypernetwork.
To train Y-Net, we follow \cite{lee2022muse} to make the encoded feature with shape $(C,H,W)$ average pooled in the spatial dimension to get a $C$ dimensional vector, and perform PCL on it. We set $a=1$ and no warmup.

\subsection{Comparisons with others}

\textbf{Quantitative comparisons on Traj++ EWTA on ETH-UCY. }
To show the effectiveness of our methods, we select  the state-of-the-art method for long-tail trajectory prediction \cite{makansi2021exposing}, classical data resampling \cite{shen2016relay} and loss reweighting \cite{cui2019class}, and a head-tail performance balancing method \cite{cao2019learning} for comparison. For long-tailed classification methods\cite{cao2019learning,shen2016relay,cui2019class}, we construct a classification head after the encoder of Traj++ EWTA to use it to classify the trajectories according to the discretization of Kalman filter errors,  same as Makansi \textit{et al.} \cite{makansi2021exposing}, and the classification loss is trained along with the prediction loss. Table \ref{tab:result} summarizes our experimental results on ETH-UCY using a best-of-20 evaluation\cite{gupta2018social}. We can see that our method stably outperforms all comparing methods on all the top $1\%-5\%$ long-tail samples. Specifically, our framework outperforms the second best method: Traj++ EWTA+contrastive \cite{makansi2021exposing} by $9.5\%$ on ADE and $8.5\%$ on FDE on the top $1\%$ hardest samples, and maintains the average ADE and FDE nearly stable. The Traj++ EWTA+reweighting \cite{cui2019class} performs best on the average ADE/FDE, but its performance gains on tailed samples are relatively little. The Traj++ EWTA+resampling \cite{shen2016relay} gets more gains on the most tailed samples, but its average ADE/FDE become much worse. Unlike simply doing resampling or loss reweighting, hypernetwork can decouple head samples and tail samples in the parameter space of decoder, therefore achieves better performances. 

\textbf{Quantitative comparisons on Traj++ EWTA on Nuscenes. } Comparison results with the previous best long-tail prediction method \cite{makansi2021exposing} on Nuscenes are in Table \ref{tab:nuscenes result}. We find out that the resampling operation in the original Traj++ EWTA does not work well with FEND, probably because of causing overfit on hypernetwork. Despite of this, as shown in Table \ref{tab:nuscenes result}, the baseline without resampling can achieve both superior long-tail and overall performances with FEND.
The performances of Traj++ EWTA and Traj++ EWTA+contrastive on both ETH-UCY and Nuscenes are tested on the provided pre-trained models of \cite{makansi2021exposing}. 

\textbf{Quantitative comparisons on Y-Net on SDD. } We also plug our module into Y-Net, the results are shown in Table \ref{tab:sdd result}.  We reproduced the results of Y-Net using the official released code of \cite{mangalam2021goals} with 42 as the random seed, since the original method does not have a fix seed. Results show that our method can achieve performance gains on both tail samples and the whole dataset.

\begin{figure}[h]
\small
\centering
  \begin{subfigure}{1.0\linewidth}
  \centering
    \includegraphics[width=0.7\linewidth]{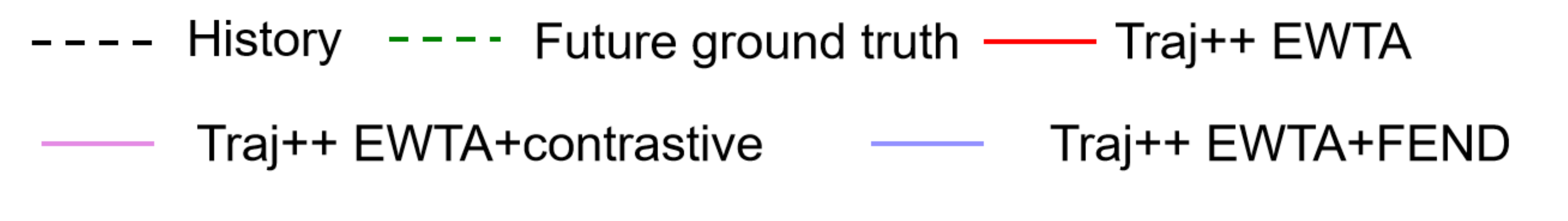}
  \end{subfigure} 
  \begin{subfigure}{0.3\linewidth}
    \fbox{\includegraphics[width=0.9\linewidth]{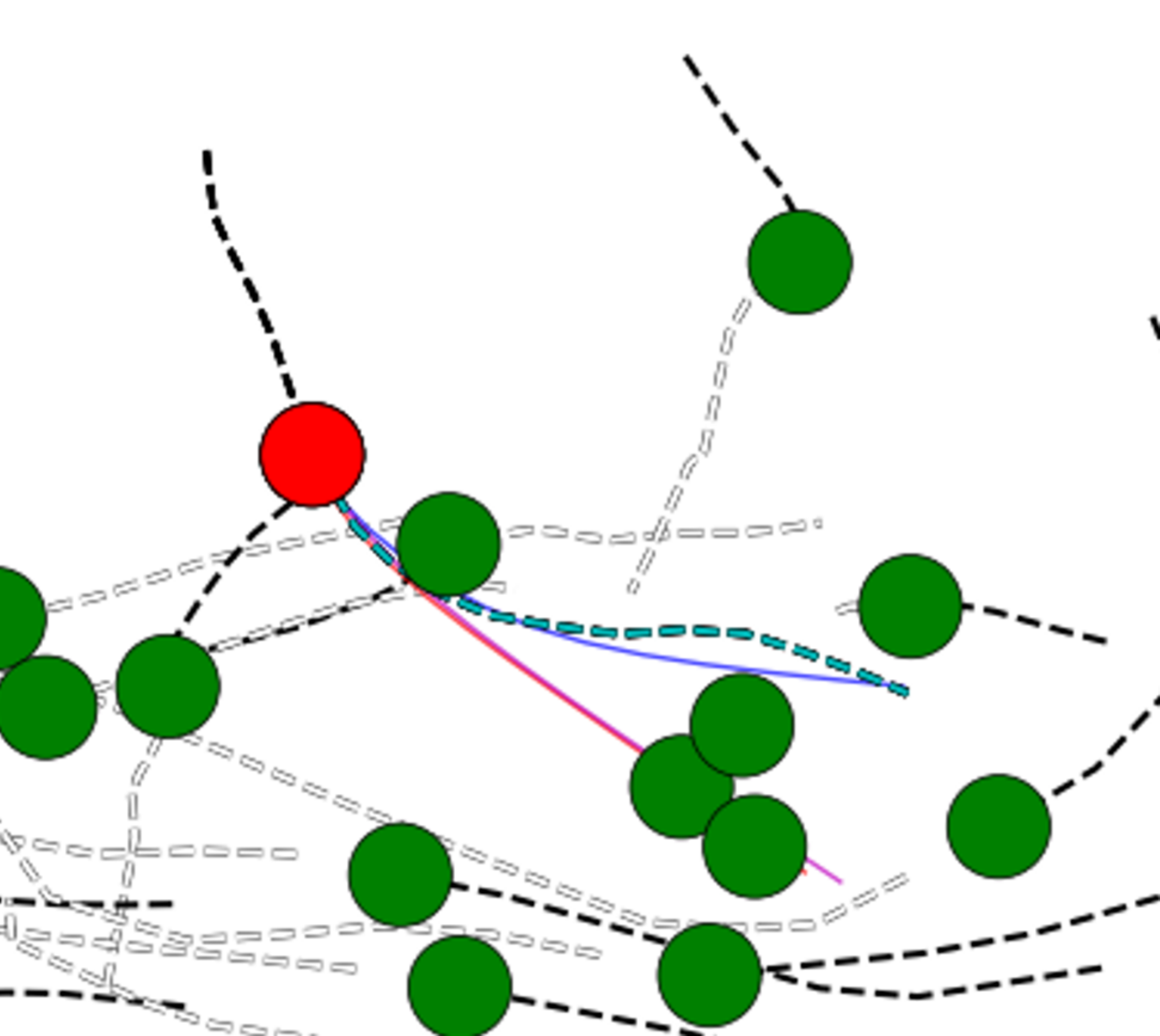}}
    \caption{}
  \end{subfigure}        
  \begin{subfigure}{0.3\linewidth}
    \fbox{\includegraphics[width=0.9\linewidth]{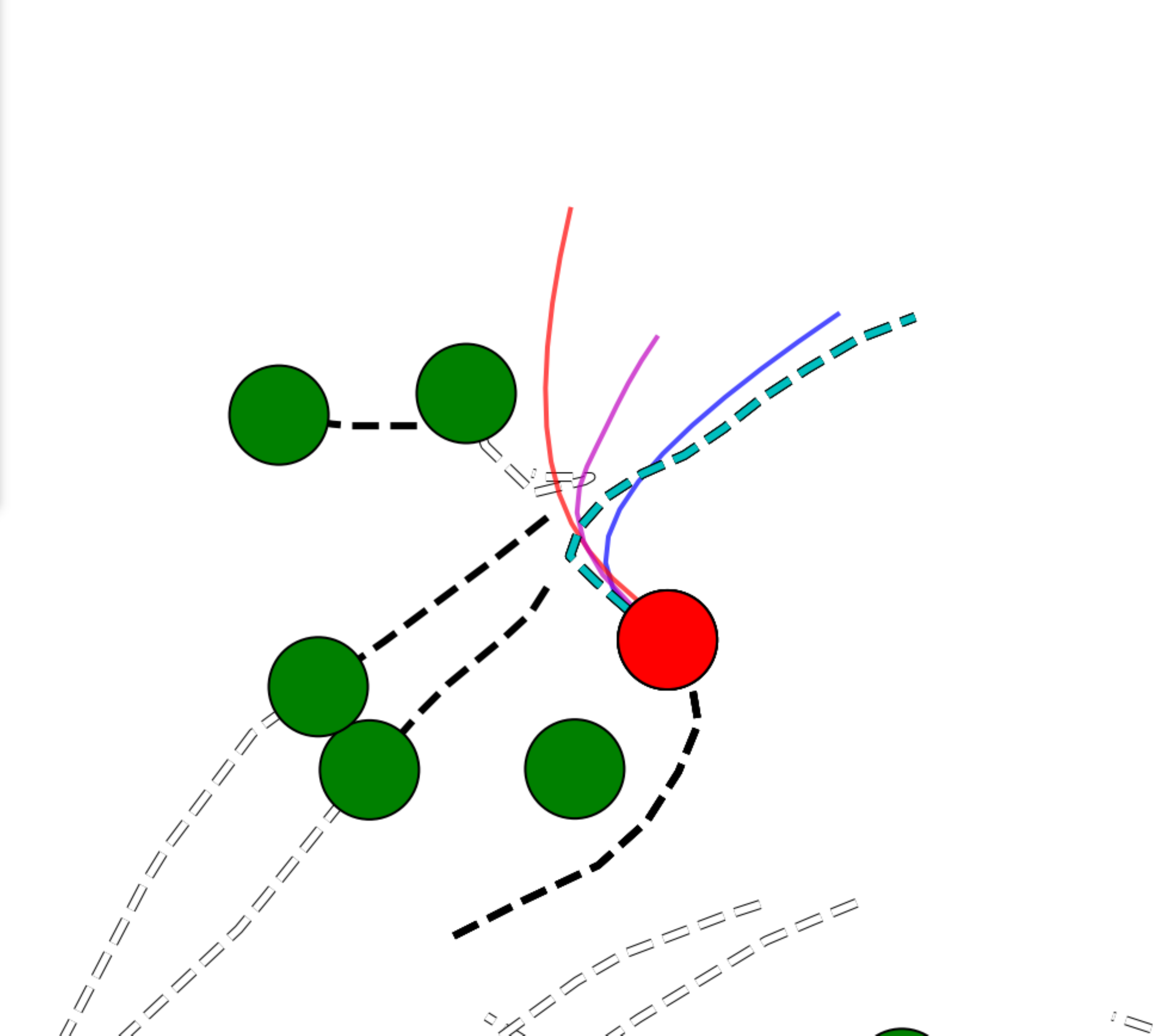}}
    \caption{}
  \end{subfigure}
    \begin{subfigure}{0.3\linewidth}
    \fbox{\includegraphics[width=0.9\linewidth]{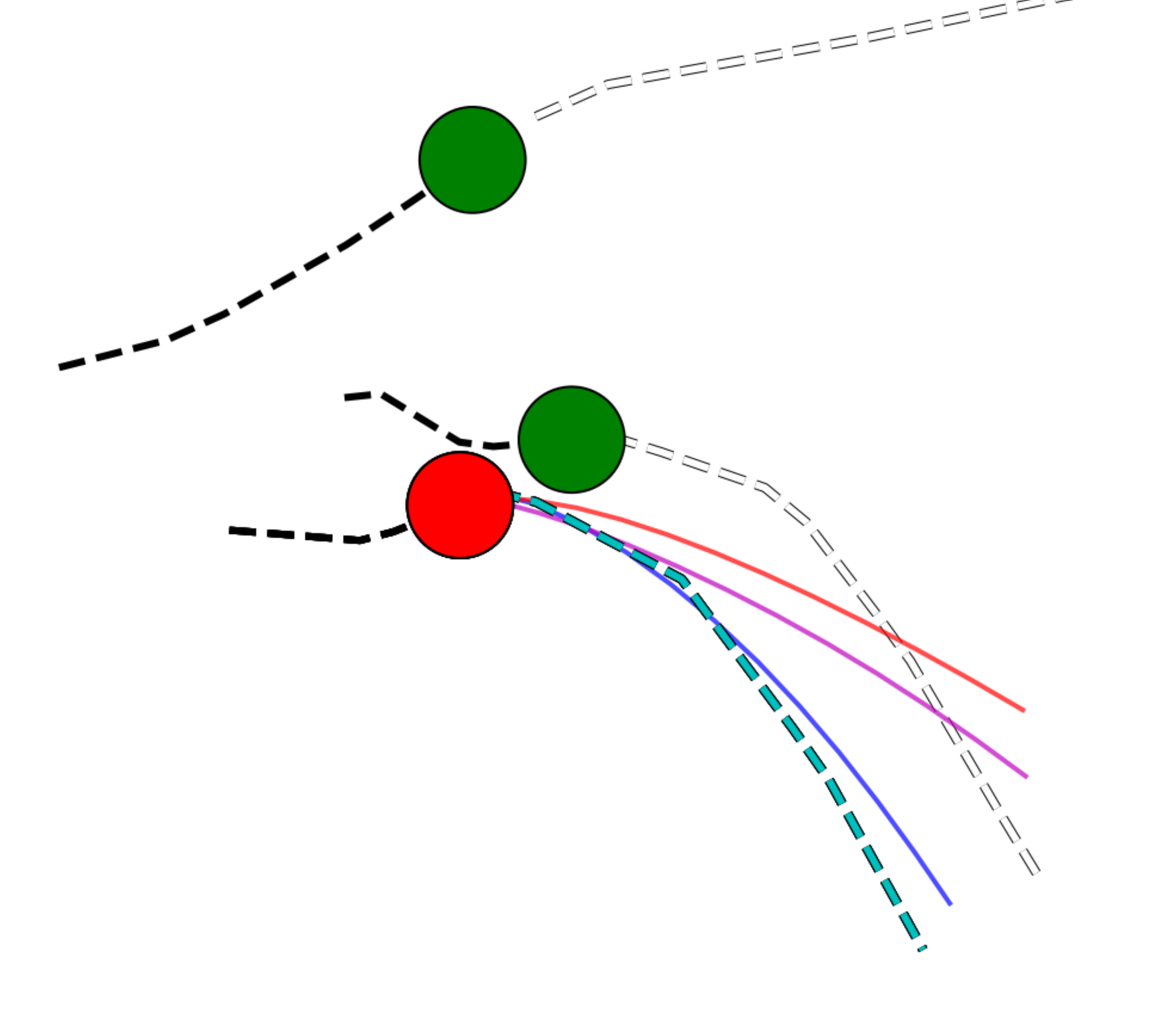}}
    \caption{}
  \end{subfigure}
      \begin{subfigure}{0.3\linewidth}
    \fbox{\includegraphics[width=0.9\linewidth]{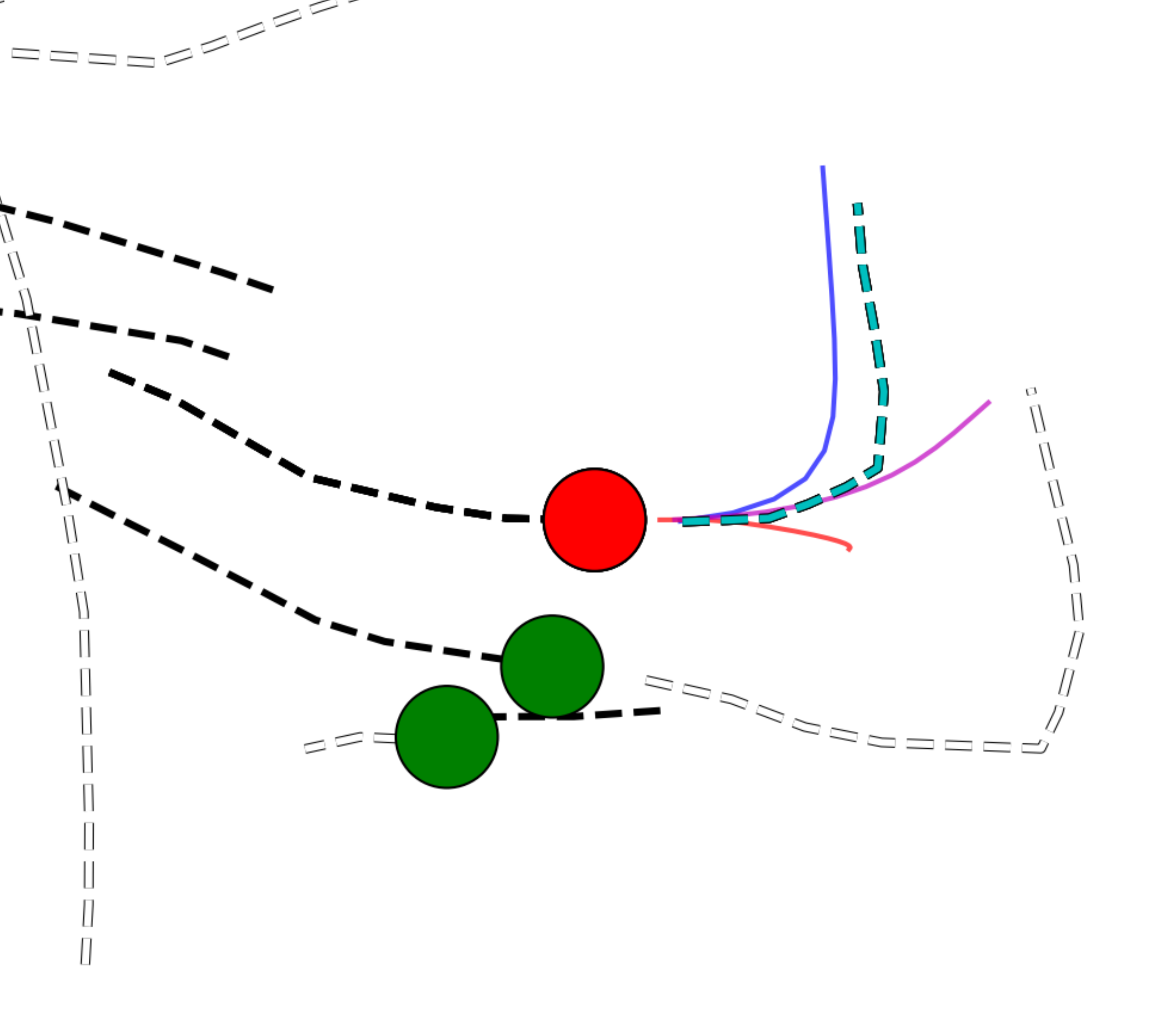}}
    \caption{}
  \end{subfigure}
      \begin{subfigure}{0.3\linewidth}
    \fbox{\includegraphics[width=0.9\linewidth]{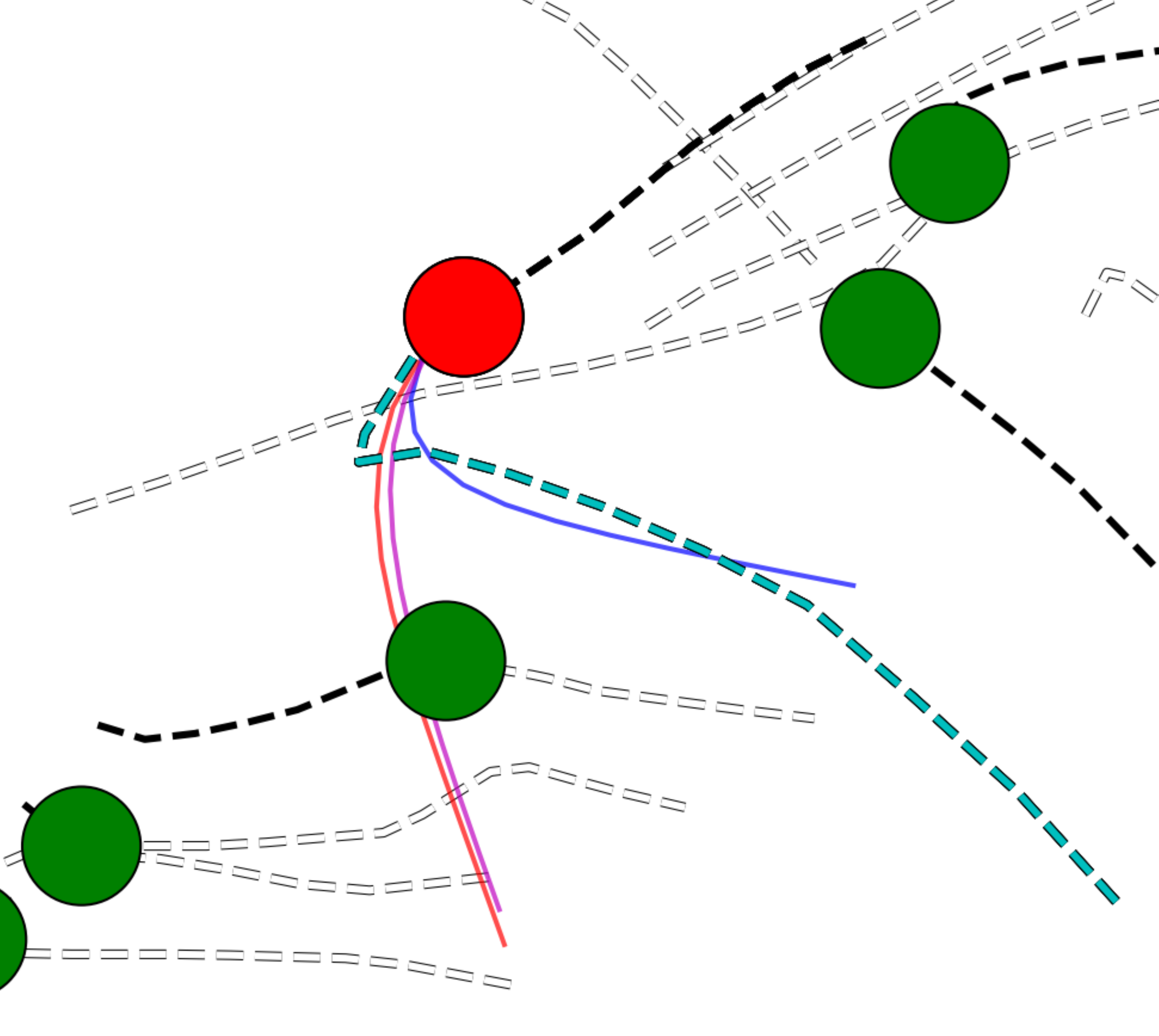}}
    \caption{}
  \end{subfigure}
      \begin{subfigure}{0.3\linewidth}
    \fbox{\includegraphics[width=0.9\linewidth]{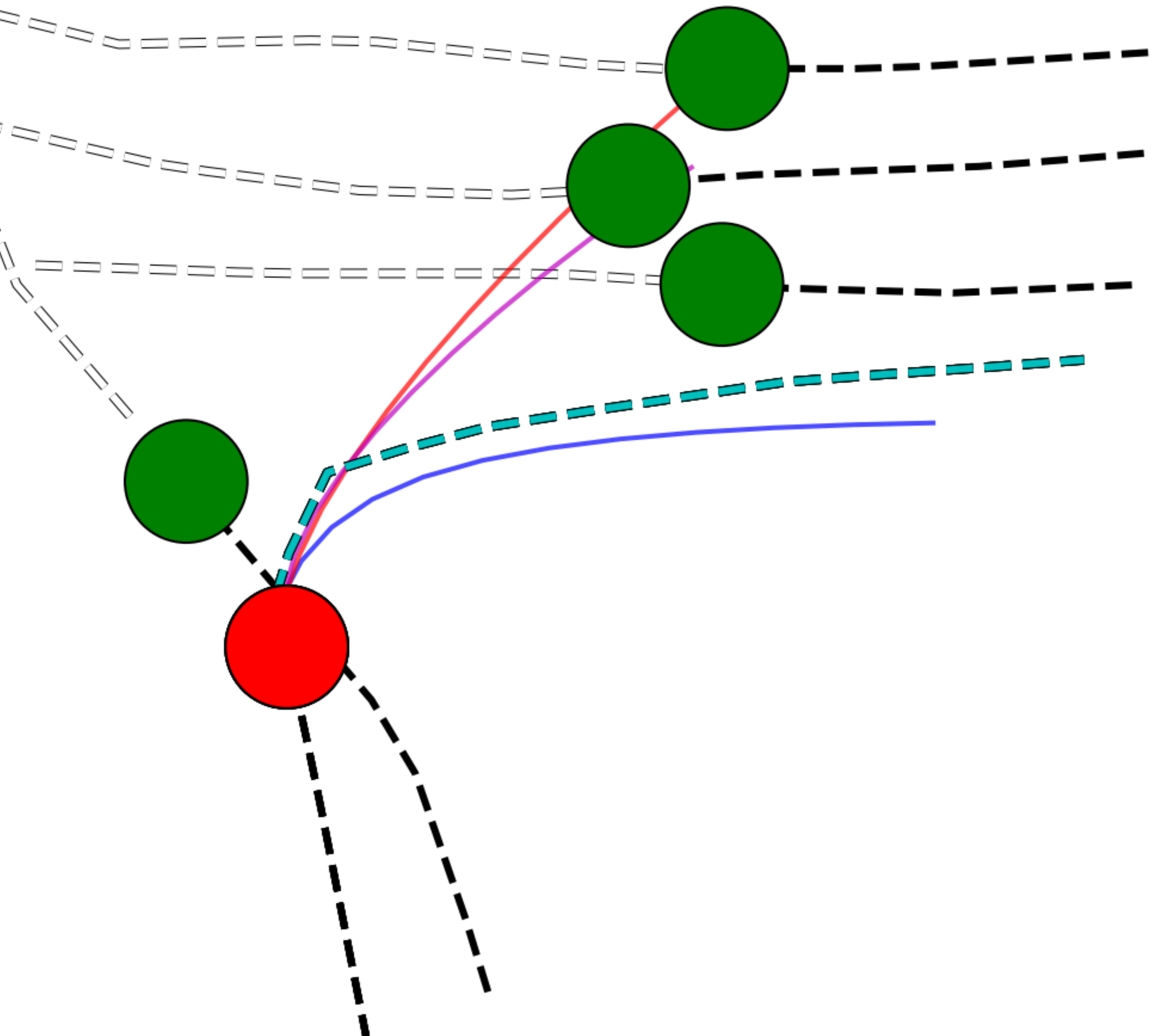}}
    \caption{}
  \end{subfigure}
\caption{Qualitative results on the ETH-UCY dataset: (a)(b) collision avoidance (c)(d) social influence of parallel walking (e)(f) crowd avoidance. The predictions are selected using a best-of-20 evaluation. }
\label{fig:qualitative}
\end{figure}

\textbf{Qualitative comparison.}
Figure \ref{fig:qualitative} shows some long-tailed hard-case studies of our method on ETH-UCY. Those cases contain some rare social interactions, and all the future trajectories in them are non-trivial to be predicted. In all those samples, our method (blue) outperforms the original Traj++ EWTA (red) and the Traj++ EWTA+contrastive (magenta), thanks to our future enhanced PCL framework for letting the prediction network better recognize different trajectory patterns and a more flexible hyper predictor.

\subsection{Ablation Study and Dicussions}

\begin{figure}[h]
\small
\centering
    \begin{subfigure}{1.0\linewidth}
  \centering
    \includegraphics[width=0.8\linewidth]{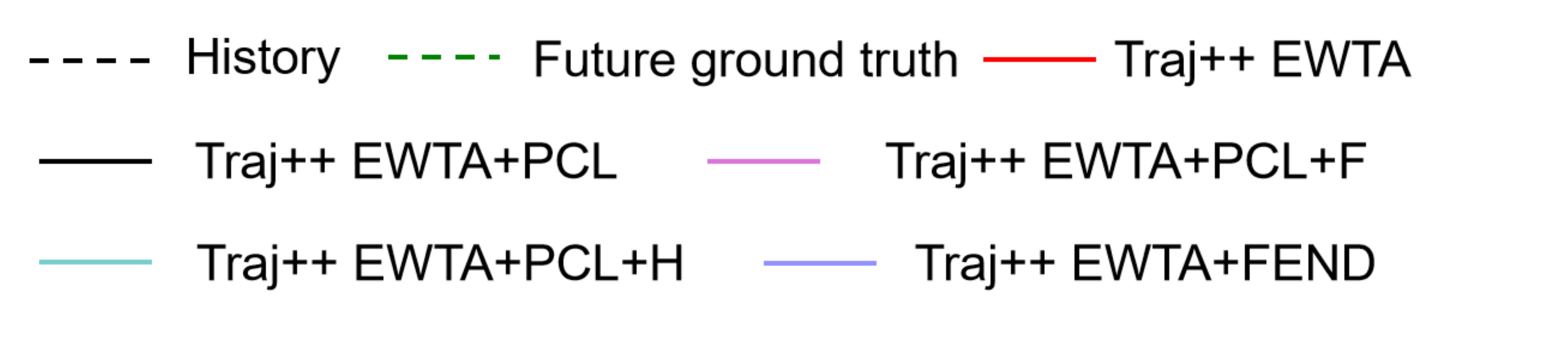}
  \end{subfigure} 
  \begin{subfigure}{0.3\linewidth}
    \fbox{\includegraphics[width=0.9\linewidth]{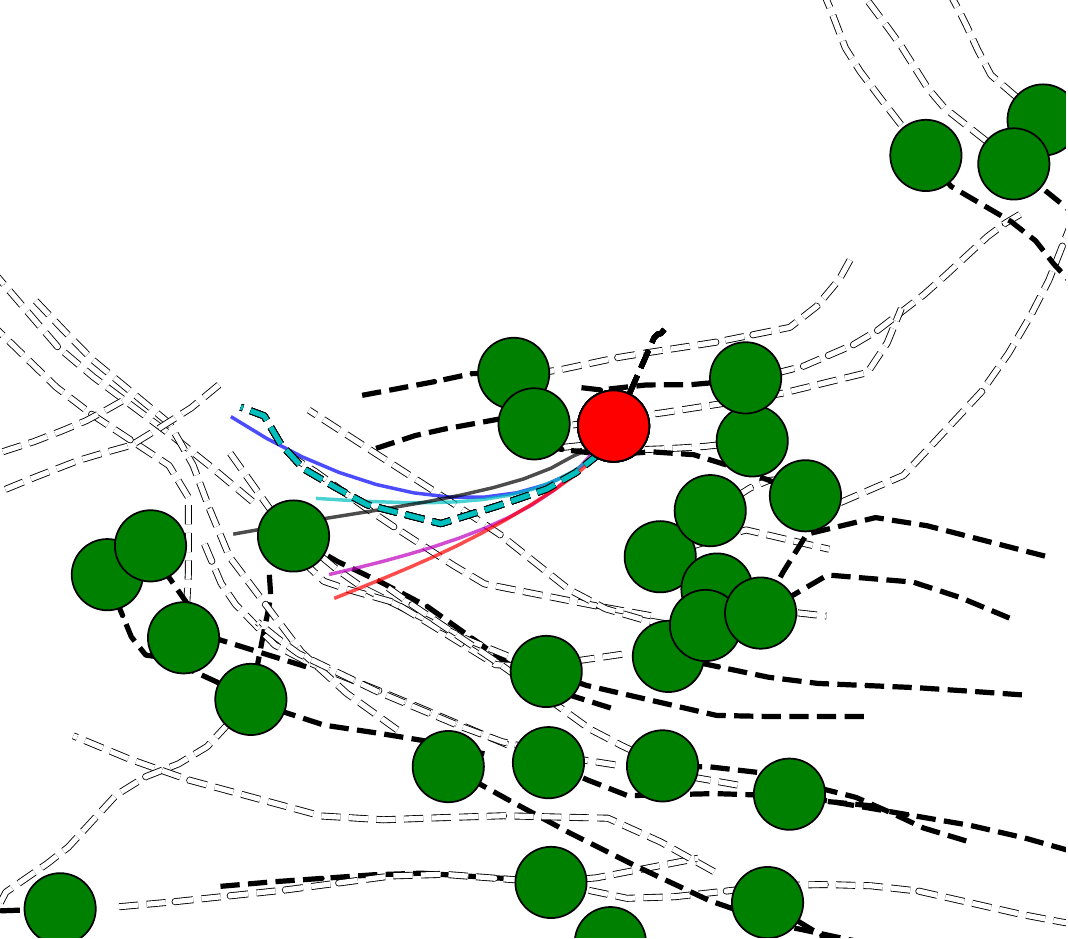}}
    \caption{}
  \end{subfigure} 
  \begin{subfigure}{0.3\linewidth}
    \fbox{\includegraphics[width=0.9\linewidth]{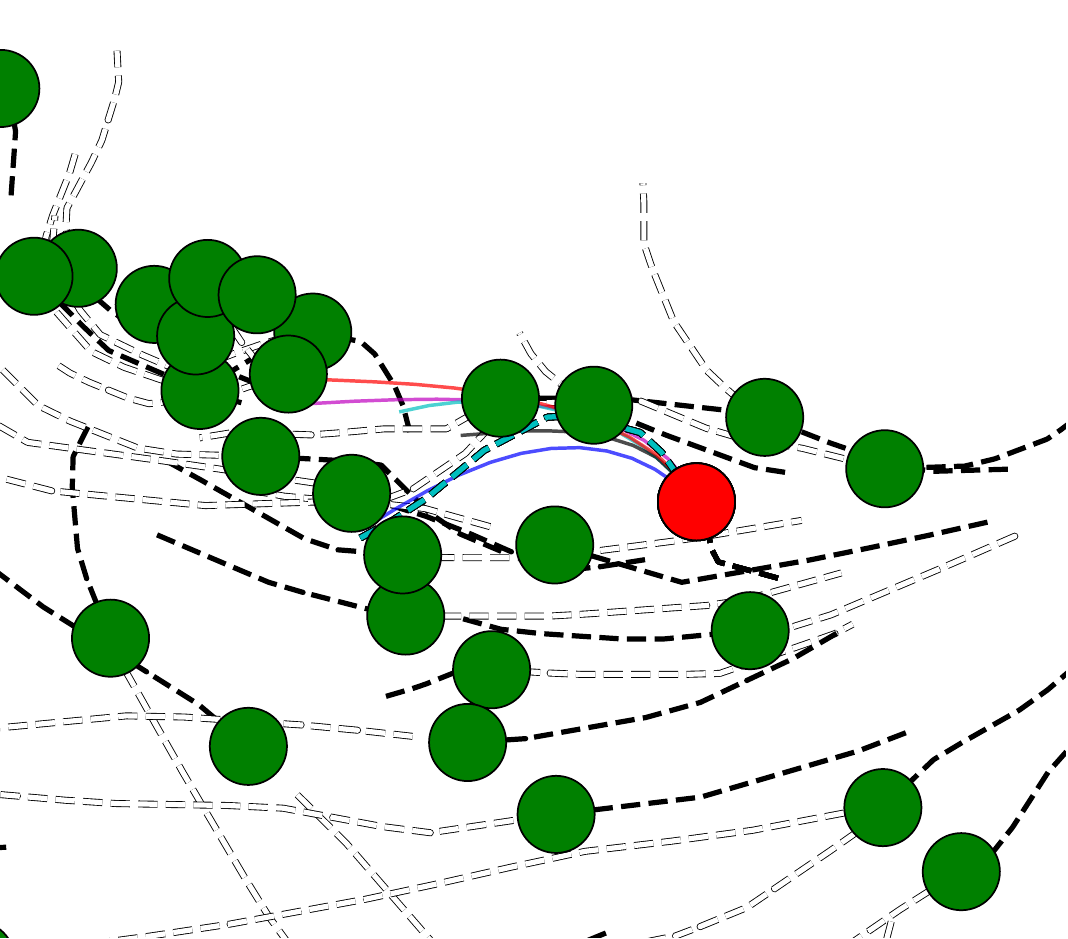}}
    \caption{}
  \end{subfigure} 
  \begin{subfigure}{0.3\linewidth}
    \fbox{\includegraphics[width=0.9\linewidth]{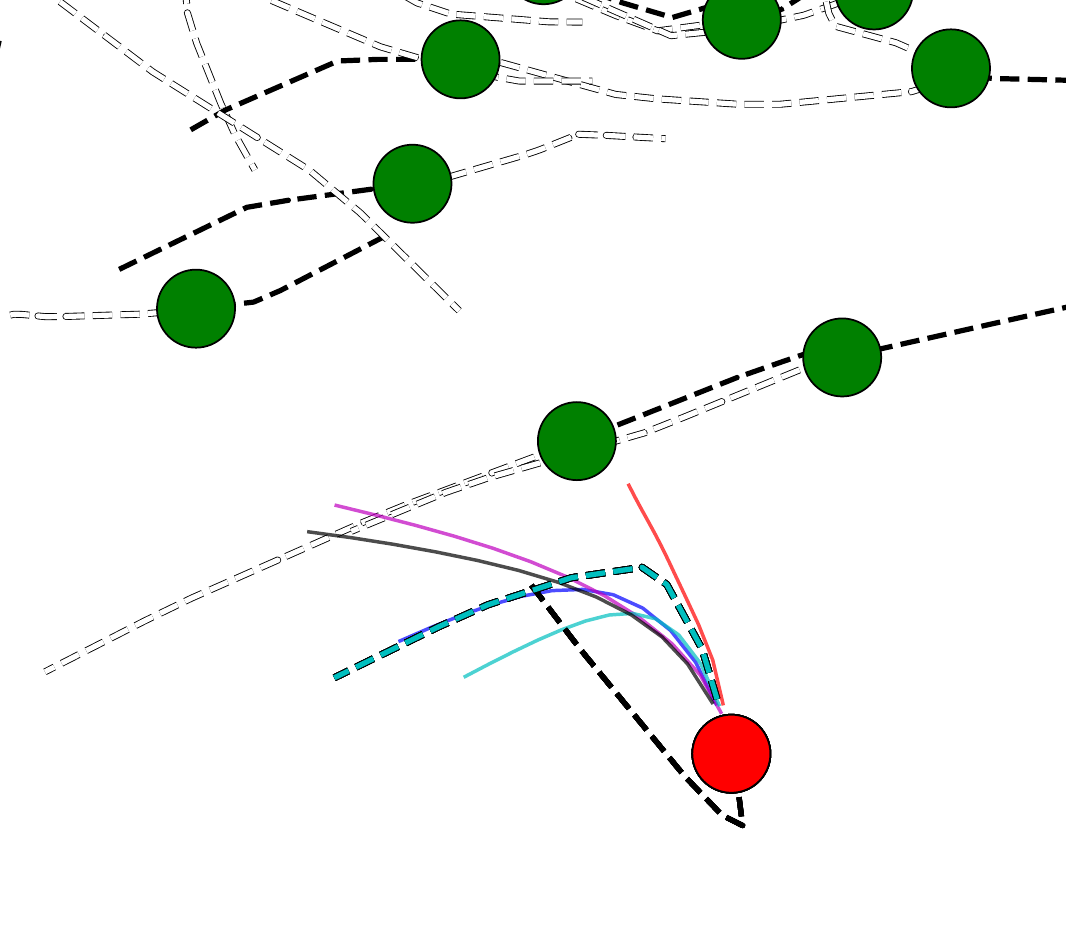}}
    \caption{}
  \end{subfigure} 
  \begin{subfigure}{0.3\linewidth}
    \fbox{\includegraphics[width=0.9\linewidth]{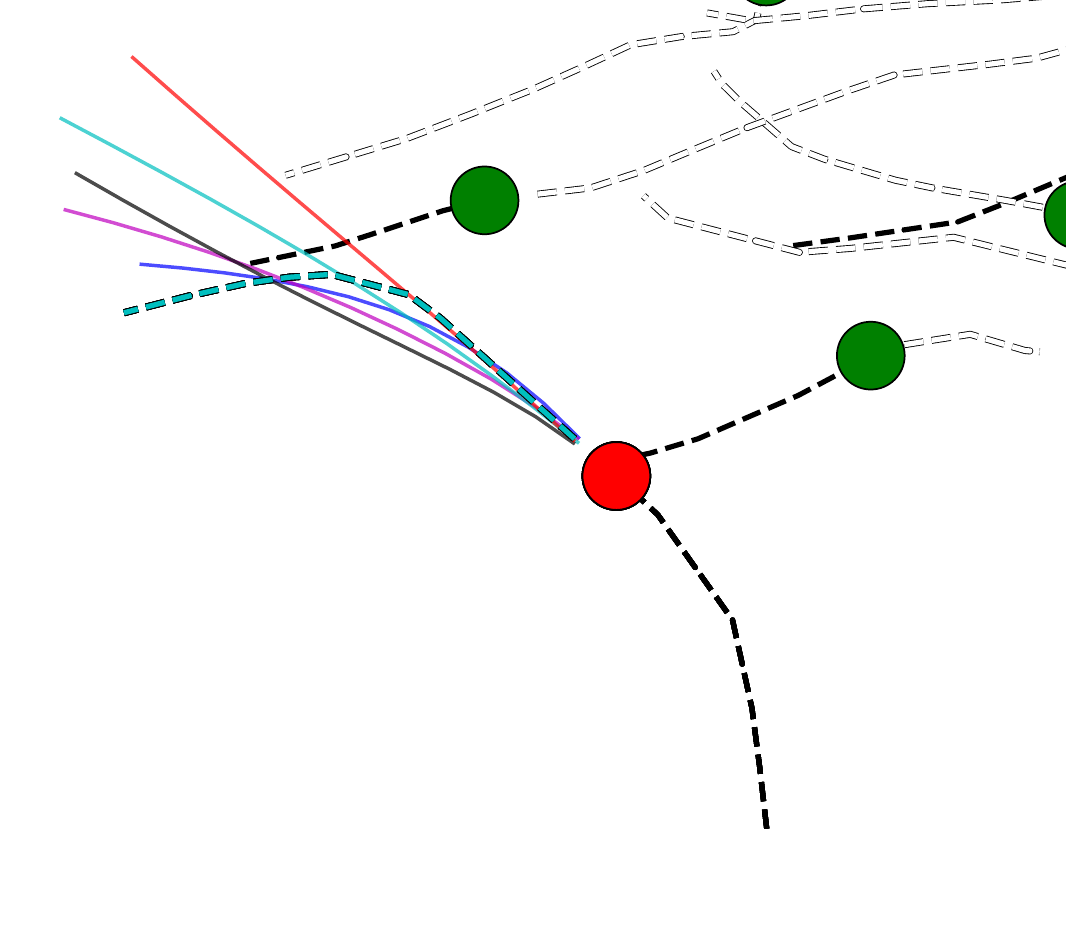}}
    \caption{}
  \end{subfigure} 
  \begin{subfigure}{0.3\linewidth}
    \fbox{\includegraphics[width=0.9\linewidth]{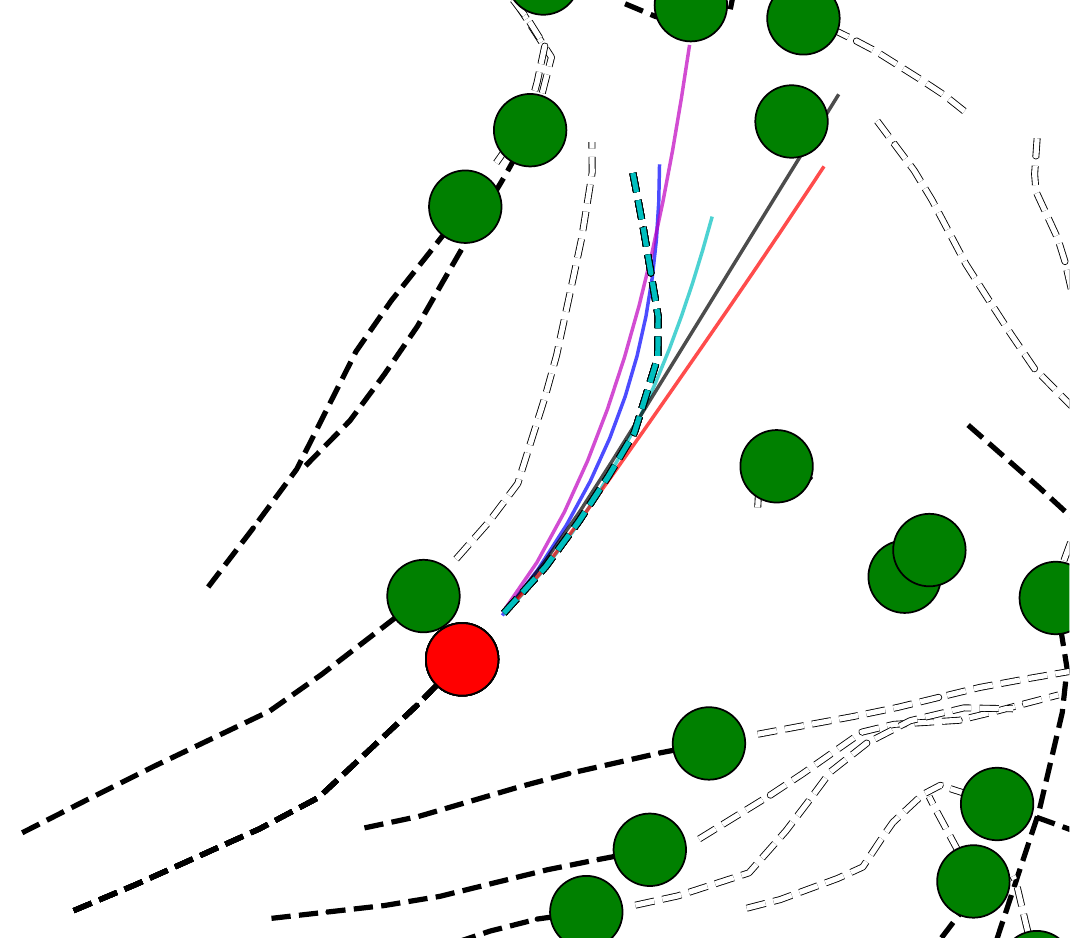}}
    \caption{}
  \end{subfigure} 
  \begin{subfigure}{0.3\linewidth}
    \fbox{\includegraphics[width=0.9\linewidth]{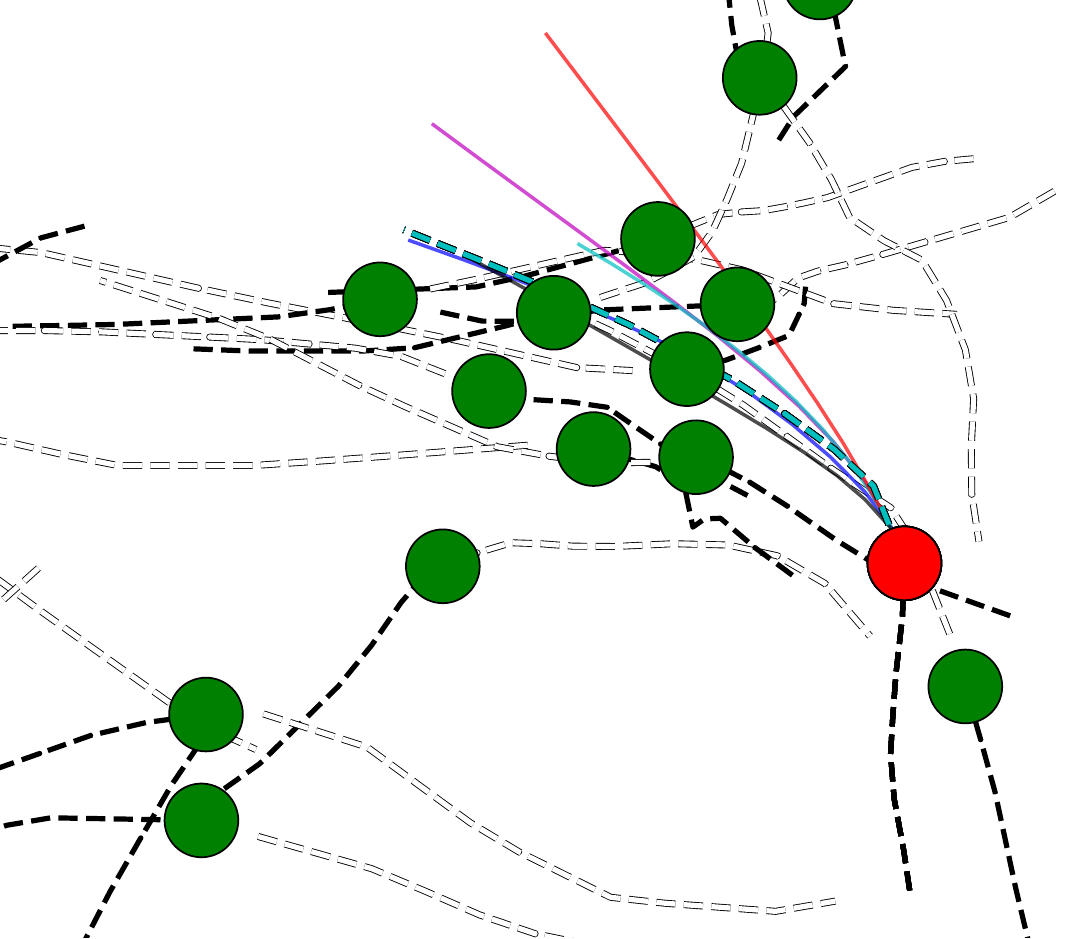}}
    \caption{}
  \end{subfigure} 
  
\caption{Qualitative results on the ETH-UCY dataset for our different model variants and the baseline Traj++ EWTA. The predictions are selected using a best-of-20 evaluation. }
\label{fig:qualitative ablation}
\end{figure}

\begin{figure*}[h]
\small
    \centering
    \begin{subfigure}{0.33\linewidth}
     \includegraphics[width=1.0\linewidth]{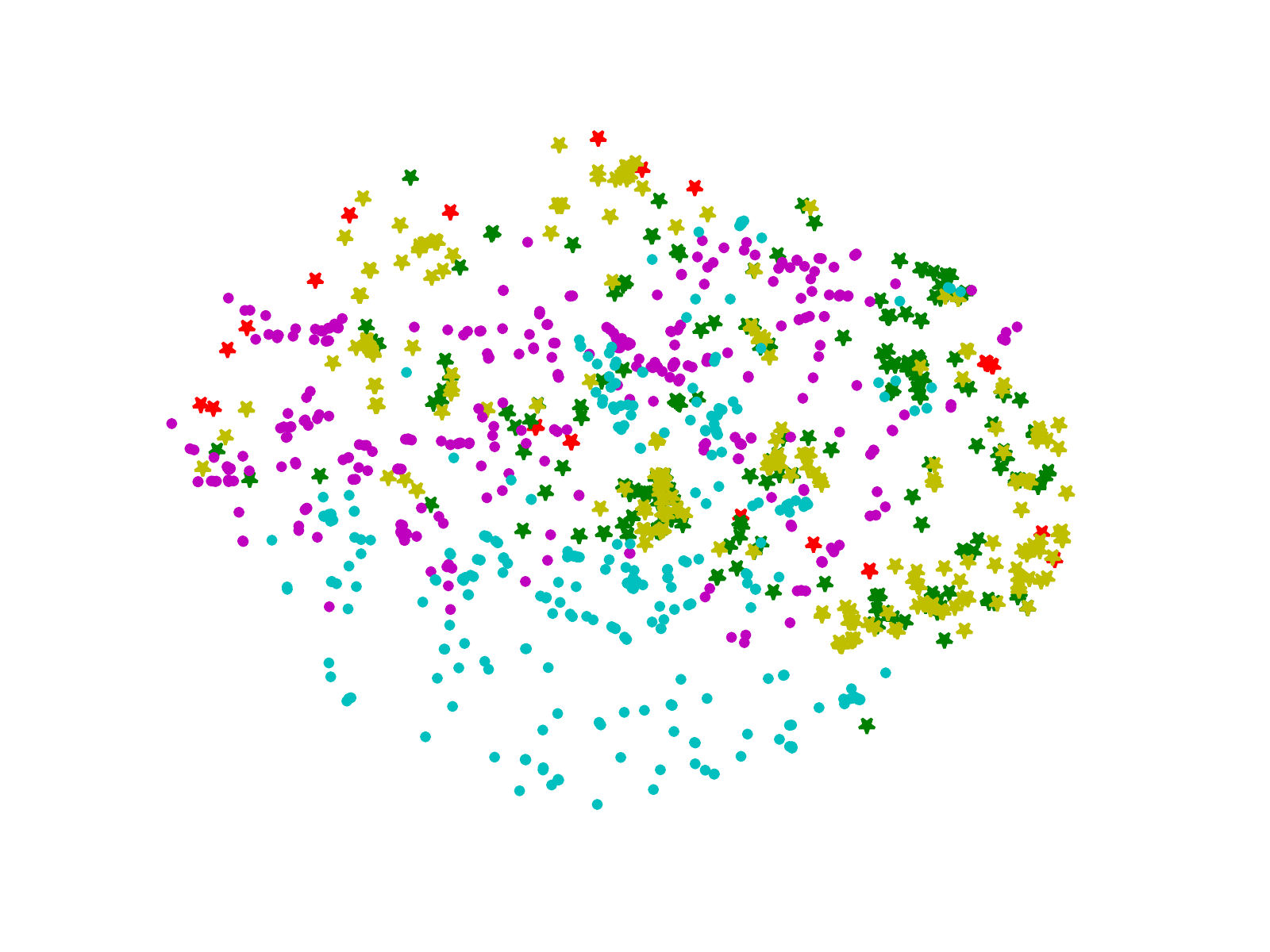}
     \caption{}
    \end{subfigure}
    \begin{subfigure}{0.33\linewidth}
     \includegraphics[width=1.0\linewidth]{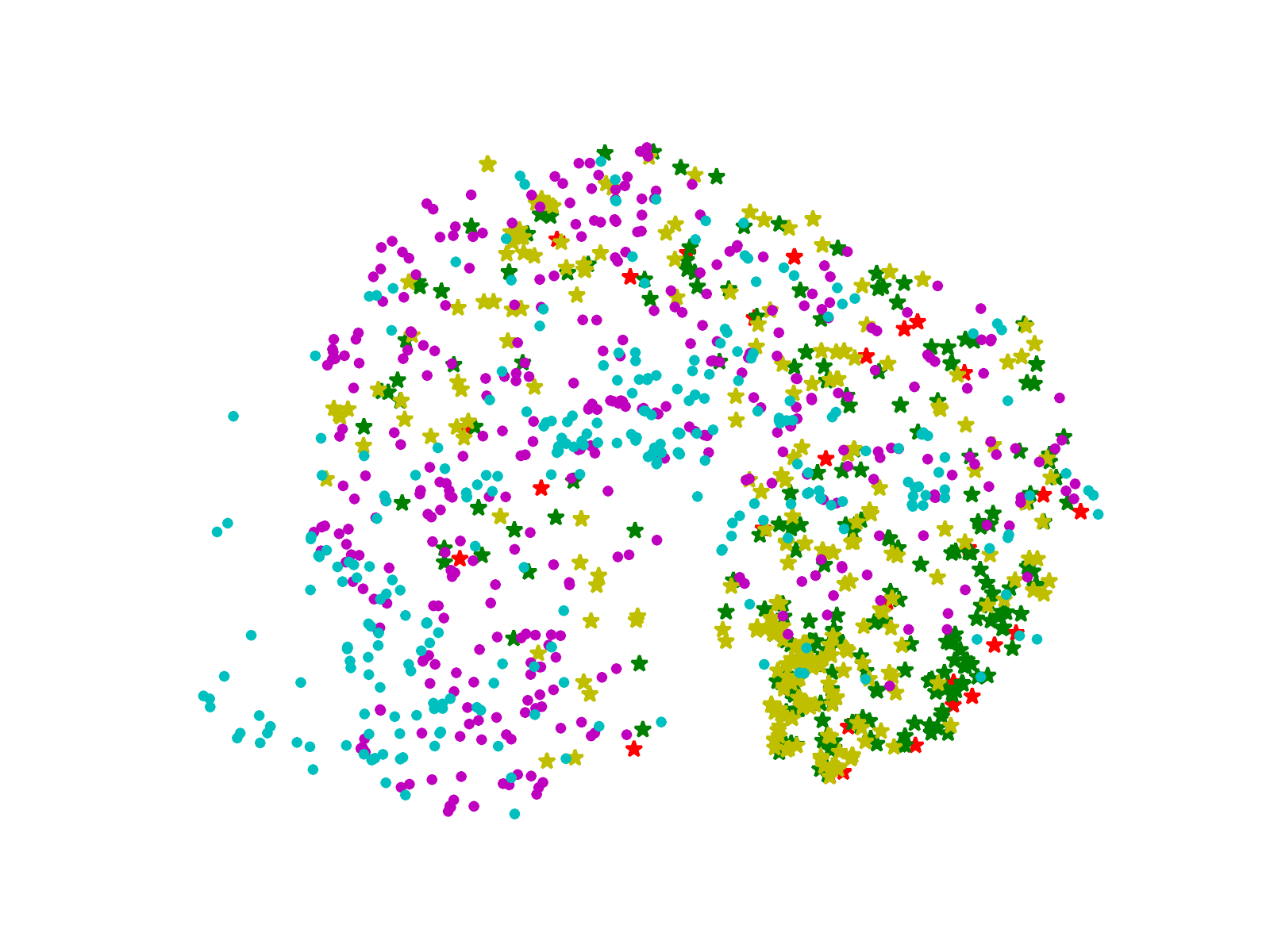}
     \caption{}
    \end{subfigure}
    \begin{subfigure}{0.33\linewidth}
     \includegraphics[width=1.0\linewidth]{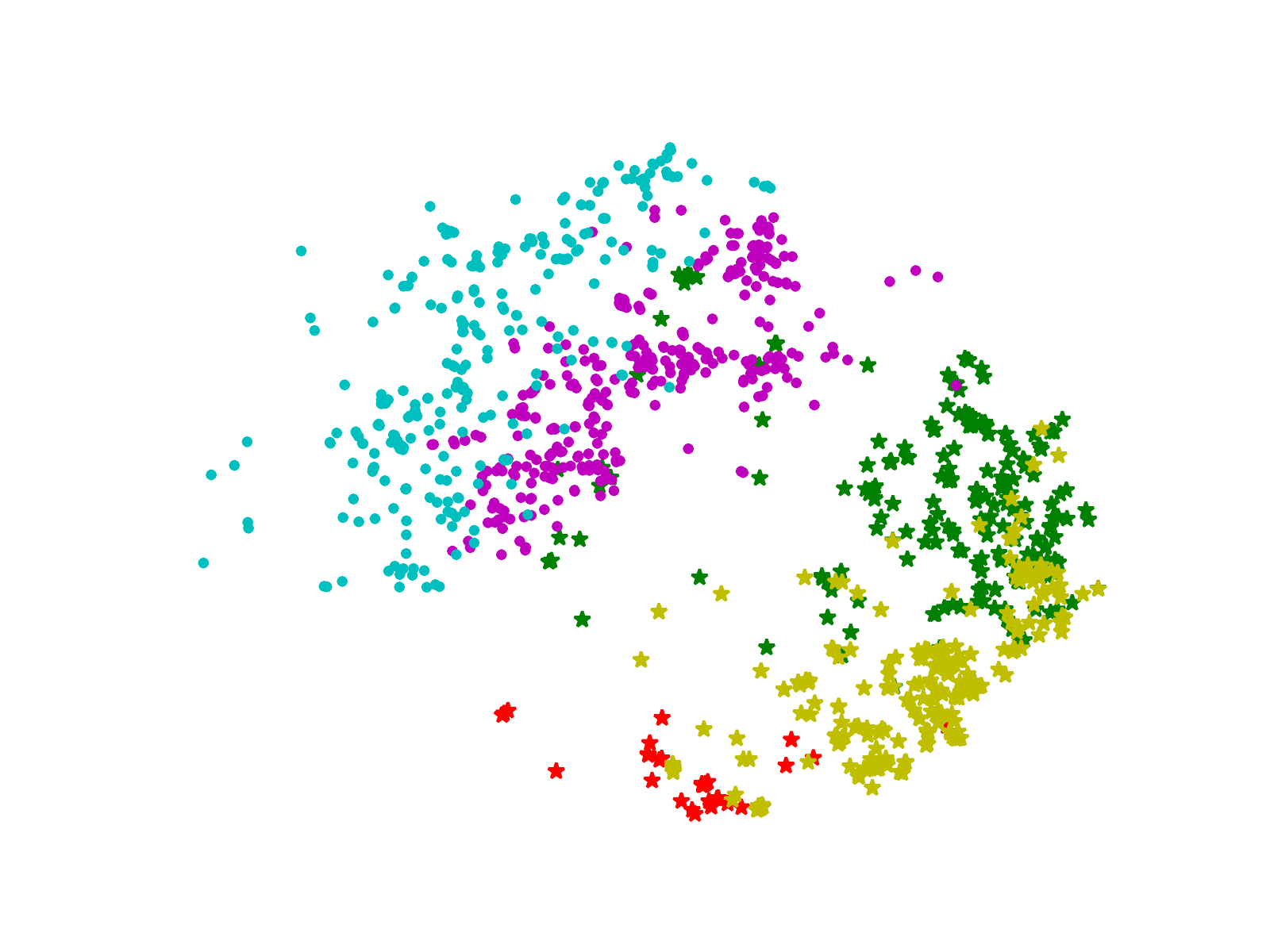}
     \caption{}
    \end{subfigure}
    \caption{TSNE results of (a)Traj++EWTA (b)Traj++ EWTA+contrastive (c)Traj++ EWTA+FEND on the univ scene. The red stars, the green stars, and the yellow stars represent clusters of three kinds of hard tailed patterns, while the magenta and cyan dots represent clusters of two kinds of easy head patterns.  We can see from the figures that our method forms a more separately clustered feature space. }
    \label{fig:tsne}
\end{figure*}

\textbf{Quantitative ablation studies.}
Results of quantitative ablation studies are shown in \cref{tab:ablation}. We can see from the results that both the future enhanced clustering and the PCL loss can contribute to the performance of the tailed samples. 
Importing the hypernetwork can also lead to a decline on the tailed FDEs. And the future enhanced PCL and the hypernetwork are compatible with each other for achieving lower tail FDEs by using both.

\textbf{Qualitative ablation studies.} Figure \ref{fig:qualitative ablation} shows some visualizations of the predict results with our different model variants. All the plotted cases are challenging, and we can see that our full model FEND stably outperforms the other variants and the baseline Traj++ EWTA. Also we can discover from the figure that all of our different model variants perform better than the baseline Traj++ EWTA.

\begin{table}[h]
\small
\centering
\begin{tabular}{llllllll}
\toprule
        $a$ & Top 1\%            & Top 2\%            & Top 3\%     & All                  \\ 
\midrule
1   & 0.97/2.48          & 0.78/2.01          & 0.69/1.72          & 0.17/0.33          \\ 
20  & 0.85/2.15          & 0.68/1.70          & 0.61/1.47          & 0.17/0.32          \\ 
50  & \textbf{0.84/2.13} & \textbf{0.68/1.68} & \textbf{0.61/1.46} & \textbf{0.17/0.32} \\ 
100 & 0.85/2.14          & 0.68/1.69          & 0.61/1.46           & 0.17/0.32          \\ 
\bottomrule
\end{tabular}
\caption{Study on the parameter sensitivity of the auxiliary loss weight $a$. Results are in the format of (minADE/minFDE) in meters.}
\label{tab: ablation a}
\end{table}

\begin{figure}[]
\small
    \centering
    \includegraphics[width=0.95\linewidth]{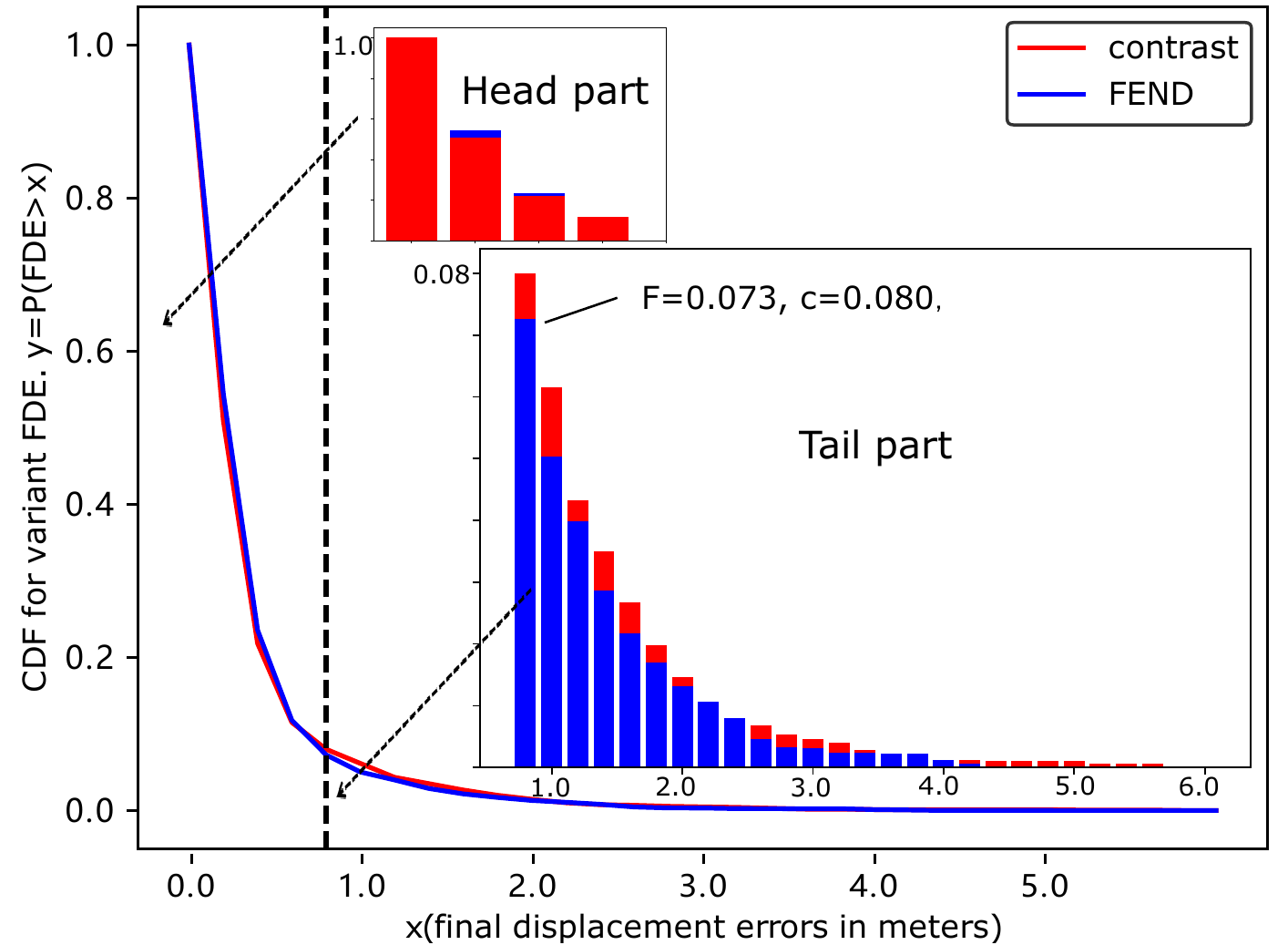}
    \caption{CDF curve and CDF bars of testing FDEs on ETH-UCY. It can be seen that our method have a shorter tail region.}
    \label{fig:cdf}
\end{figure}

\textbf{Parameter sensitivity study.}
Table \ref{tab: ablation a} shows the parameter sensitivity study of PCL loss weight $a$. We can see that setting $a=50$ initially will be the best choice. Other parameter sensitivity studies are provided in supplementaries.

\textbf{Shaped feature embedding space.}
Figure \ref{fig:tsne} shows the TSNE results of the feature space of our method and two comparing methods, with two head patterns and three tailed patterns. We can see from the figure that our future enhanced PCL method can decently separate the tail patterns and the head patterns, while there is still some overlap between the heads and the tails in the feature space of Traj++ EWTA and Traj++ EWTA+contrastive. Also, we can see from \cref{fig:tsne} that our method can form different clusters for different tailed patterns, while in the feature space of Traj++ EWTA+contrastive, all the samples of the three tail patterns are pushed together, as in \cref{sec:related} discussed.

\textbf{FDE distribution bars.} To illustrate the distribution of the prediction errors across the dataset more clearly, We plot the cumulative distribution function (CDF) curve of FDEs, and the CDF bars of head and tail regions on ETH-UCY in
Figure \ref{fig:cdf} versus the second-best: Traj++ EWTA+contrastive \cite{makansi2021exposing}. The CDF is averaged across the five scenes.

\textbf{Limitations.} The performances on the head samples are slightly dropped, which can been seen in Figure \ref{fig:cdf} and Table \ref{tab:result} \ref{tab:nuscenes result} \ref{tab:sdd result}. We leave it as future works. In most experiments we use the minADE/FDE as the prediction evaluation protocols. 
There are many better metrics such as the Negative Log-Likelihood (NLL) \cite{hug2020introducing,ivanovic2019trajectron,bhattacharyya2018accurate} or those which take scene-compliance or socially acceptable prediction into account \cite{kothari2021human,ivanovic2022injecting}.
The results of another evaluation protocol: FDE NLL are in supplementaries.

\textbf{Discussion about single agent clustering.} We use single agent full trajectory features for clustering, similar to other works using single trajectories to cluster or retrieve \cite{sun2021three,zhao2021you}.
In our experiment we find out that the information in single agent trajectories can already lead to good performances. 
We believe that it is a promising future direction to include social features into the clustering process.

\section{Conclusion}
In this paper, we propose a future enhanced contrastive feature space shaping method and a distribution-aware hyper decoder for long-tailed trajectory prediction. Quantitive and qualitative experiment results show that our method can outperform state-of-the-art long-tail prediction methods on the challenging tailed samples, while maintaining the averaged performance on the whole datasets. Our method can be generally plugged into many strong prediction networks.

\section*{Acknowledgement}
This work was supported by NSFC Projects (No. 62036008) and STI 2030—Major Projects (No. 2021ZD0201300). 

{\small
\bibliographystyle{ieee_fullname}
\bibliography{egbib}
}

\clearpage
\appendix

\section{Visualization Of Different Prototypes}

In our framework FEND, we first do trajectory clustering to get different trajectory prototypes. To make the trajectory clustering more focused on the motion patterns of trajectories, we normalize the trajectories before clustering. Specifically, we translate the final observed trajectory points to the origin of the coordinate system. After that, we rotate the trajectories to make the velocity at the last observed time along the coordinate axis. Besides, we use the same trajectory normalization as Trajectron++ \cite{salzmann2020trajectron++}, which make the distribution of trajectory lengths the same across scenes. After the normalization of Trajectron++, the lengths of history trajectories will be in $[0,2]$ and shorter than future trajectories.

\begin{figure}[h]
\centering
\begin{subfigure}{0.30\linewidth}
\centering
\includegraphics[width=1.0\linewidth]{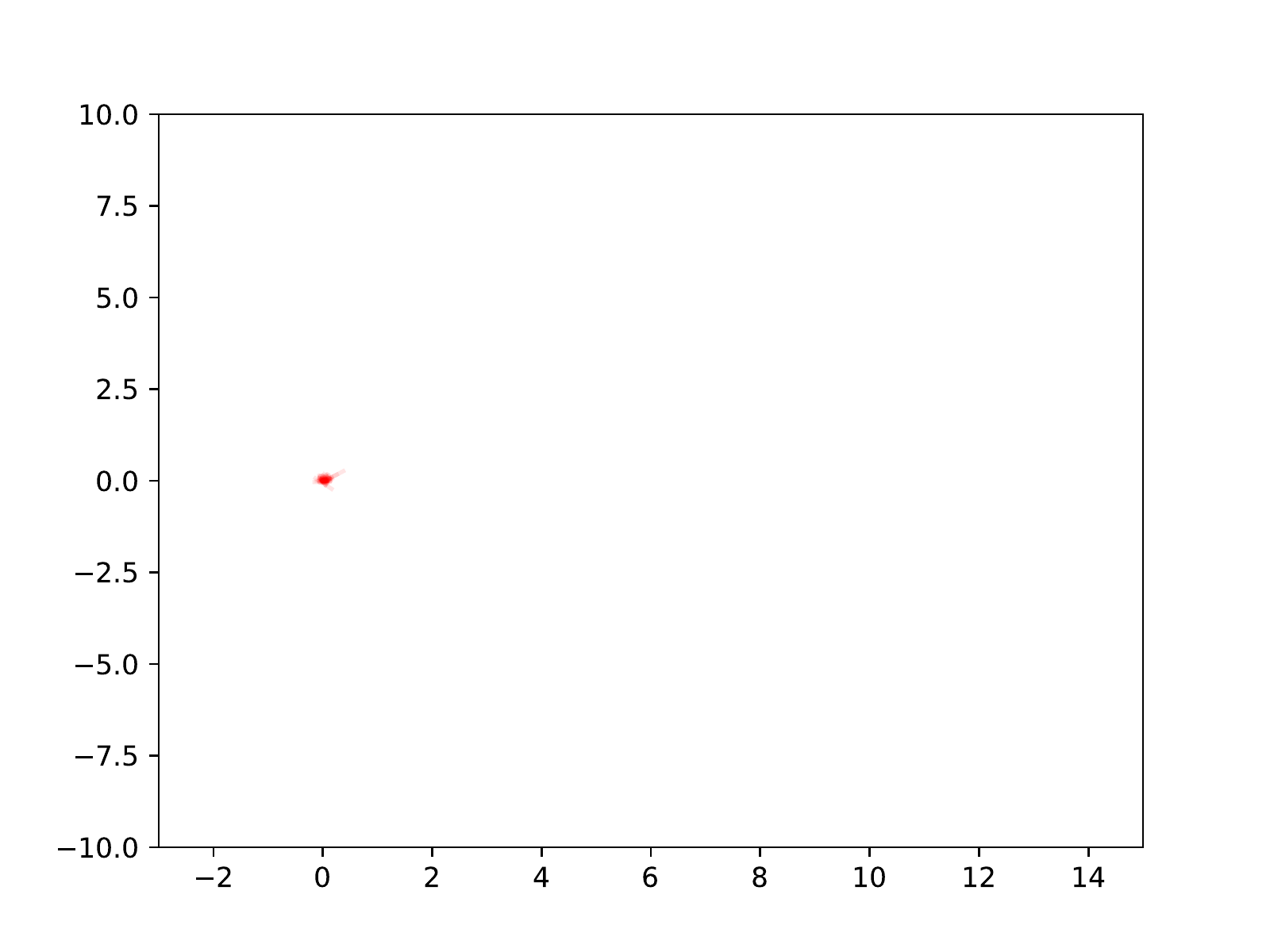}
\caption{}
\end{subfigure}
\begin{subfigure}{0.30\linewidth}
\centering
\includegraphics[width=1.0\linewidth]{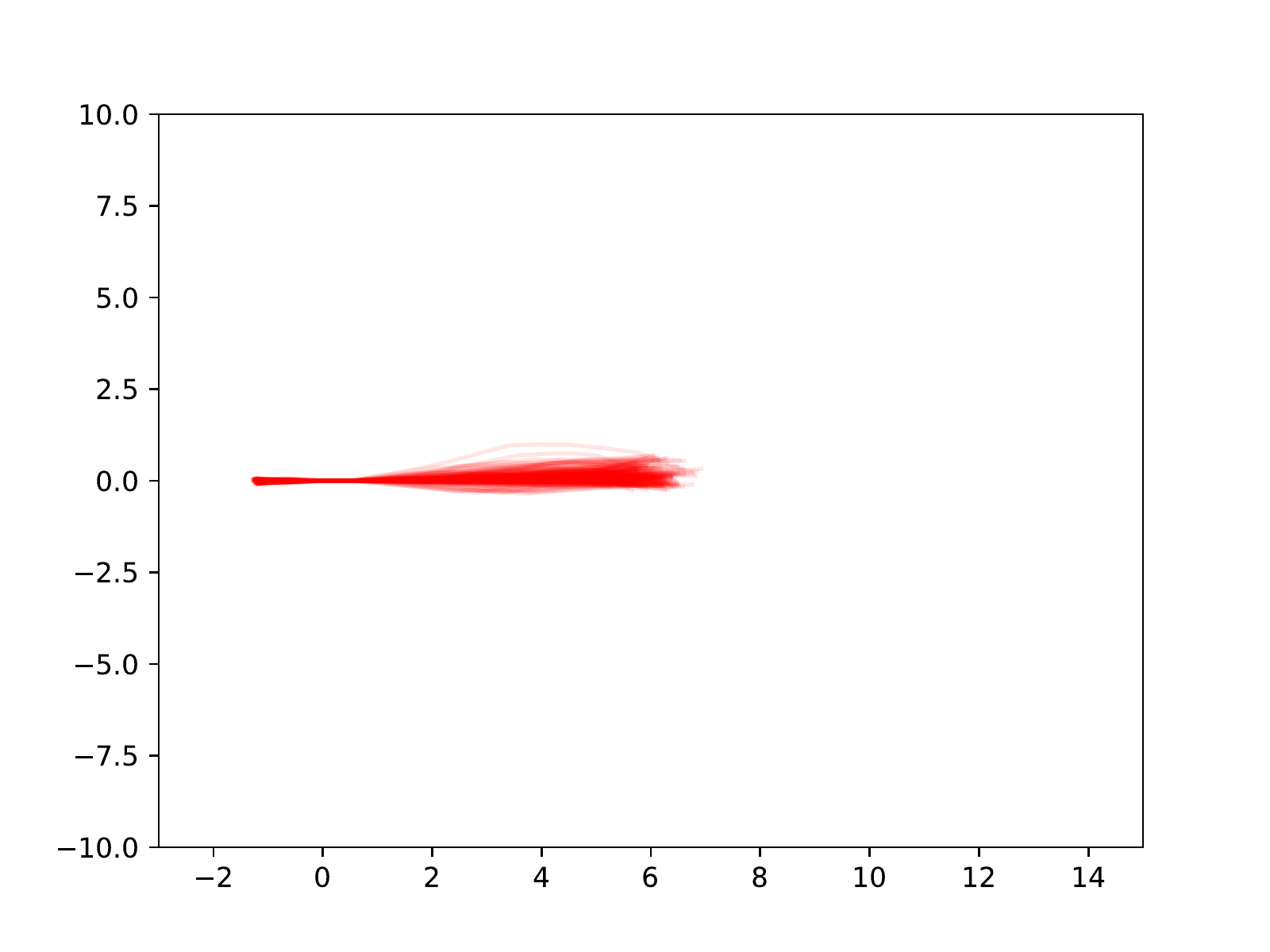}
\caption{}
\end{subfigure}
\begin{subfigure}{0.30\linewidth}
\centering
\includegraphics[width=1.0\linewidth]{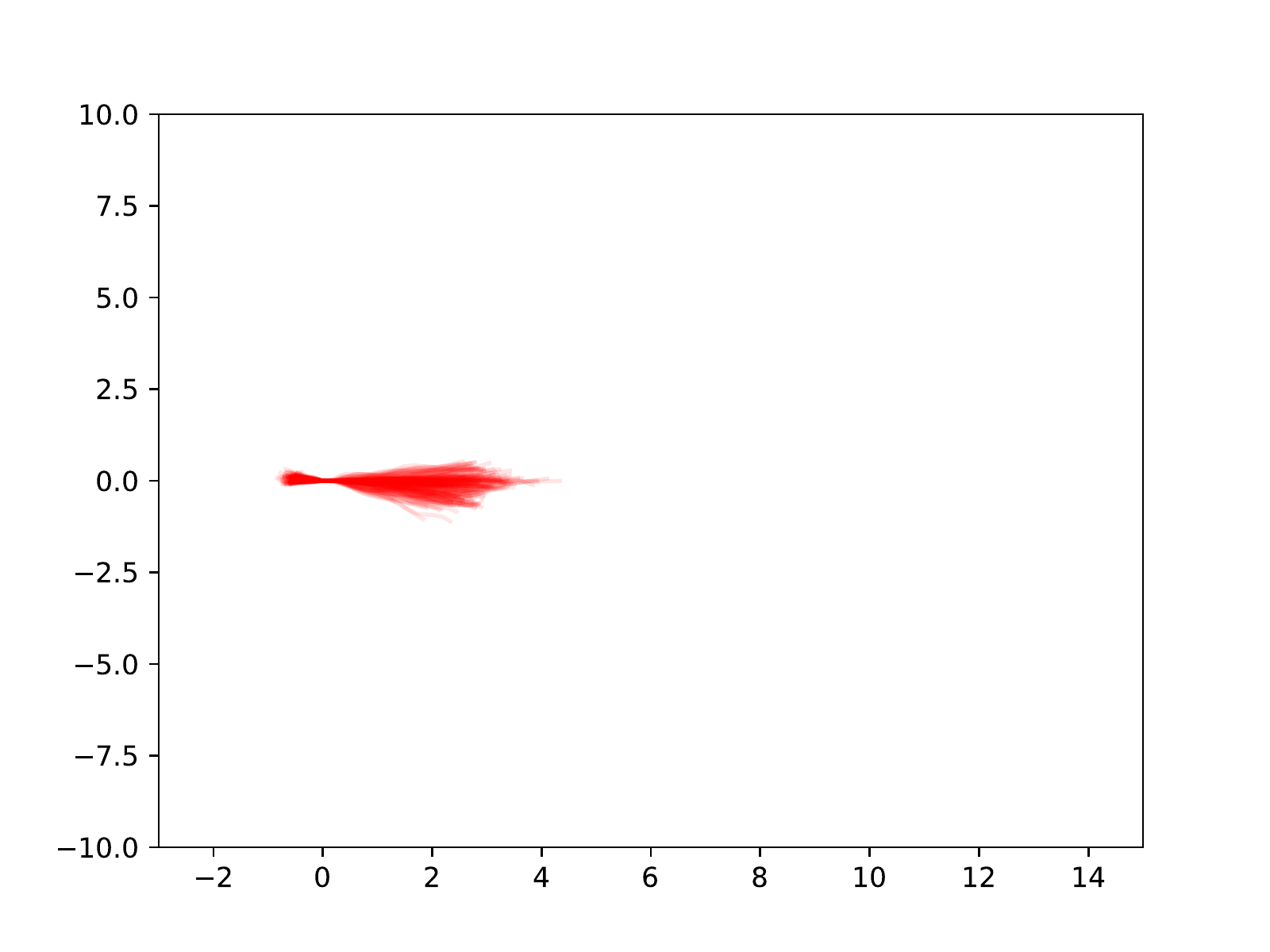}
\caption{}
\end{subfigure}
\caption{Trajectories in different head prototype clusters.}
\label{fig:head}
\end{figure}

\begin{figure}[h]
\centering
\begin{subfigure}{0.30\linewidth}
\centering
\includegraphics[width=1.0\linewidth]{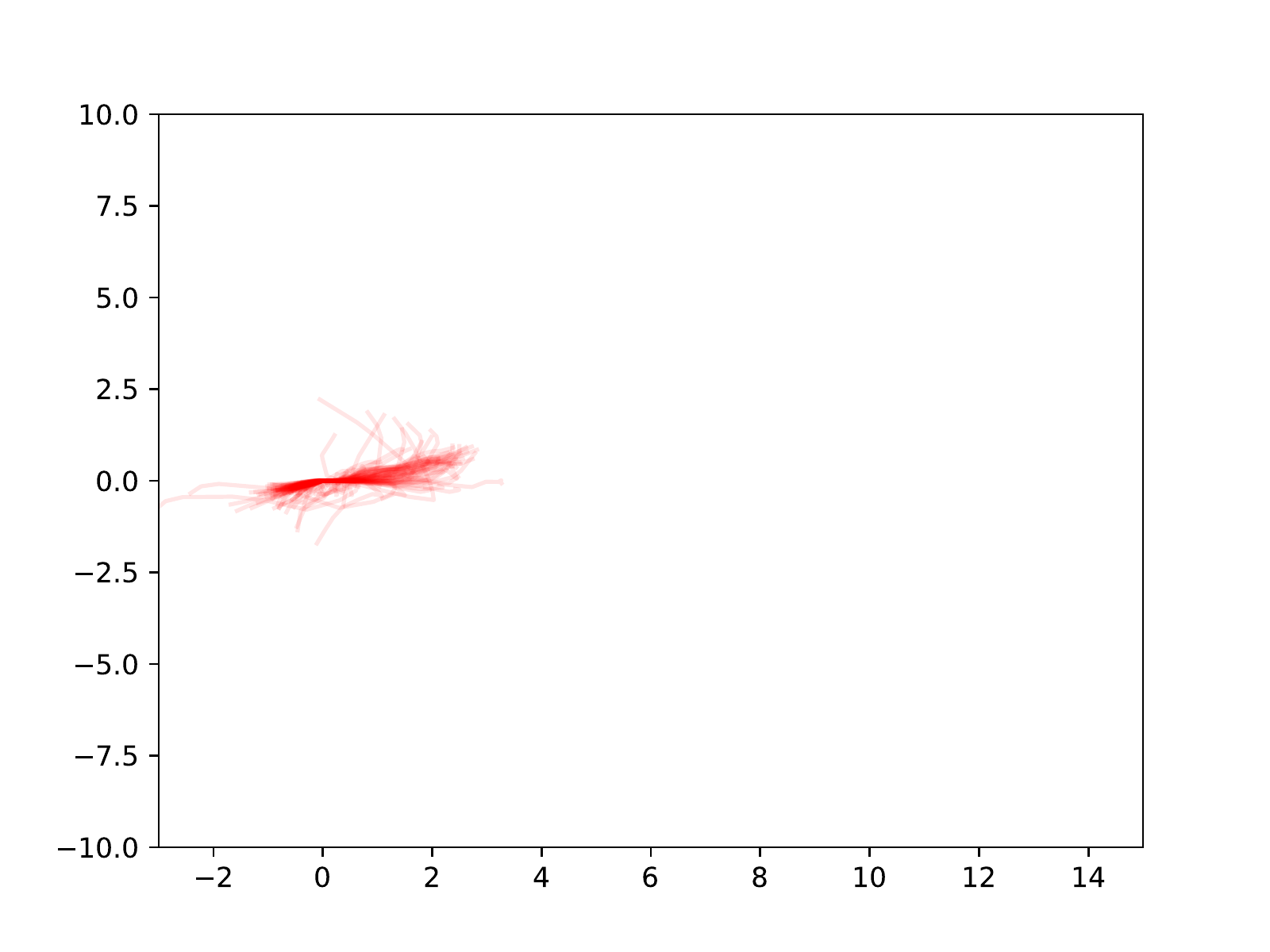}
\caption{}
\end{subfigure}
\begin{subfigure}{0.3\linewidth}
\centering
\includegraphics[width=1.0\linewidth]{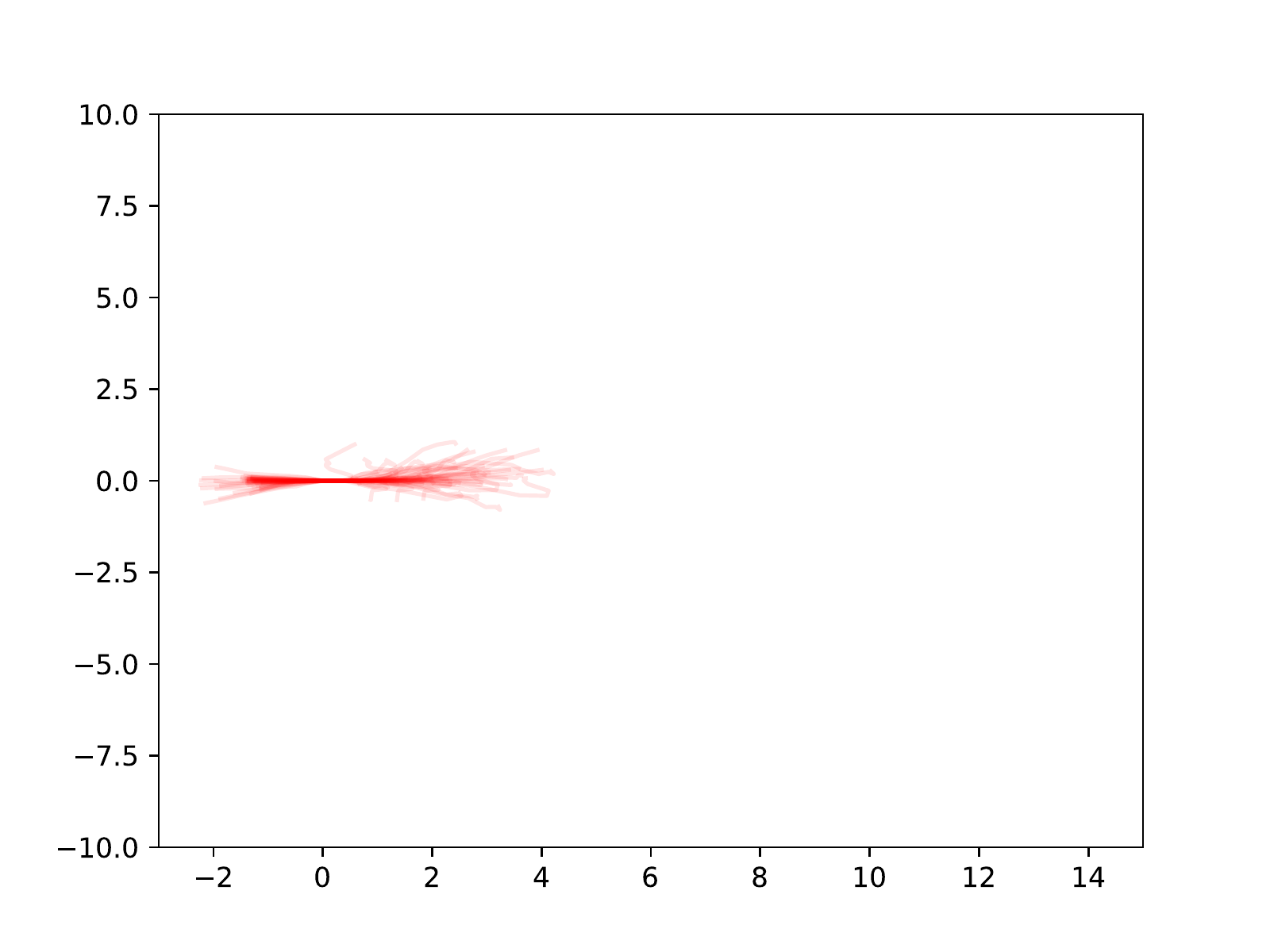}
\caption{}
\end{subfigure}
\begin{subfigure}{0.3\linewidth}
\centering
\includegraphics[width=1.0\linewidth]{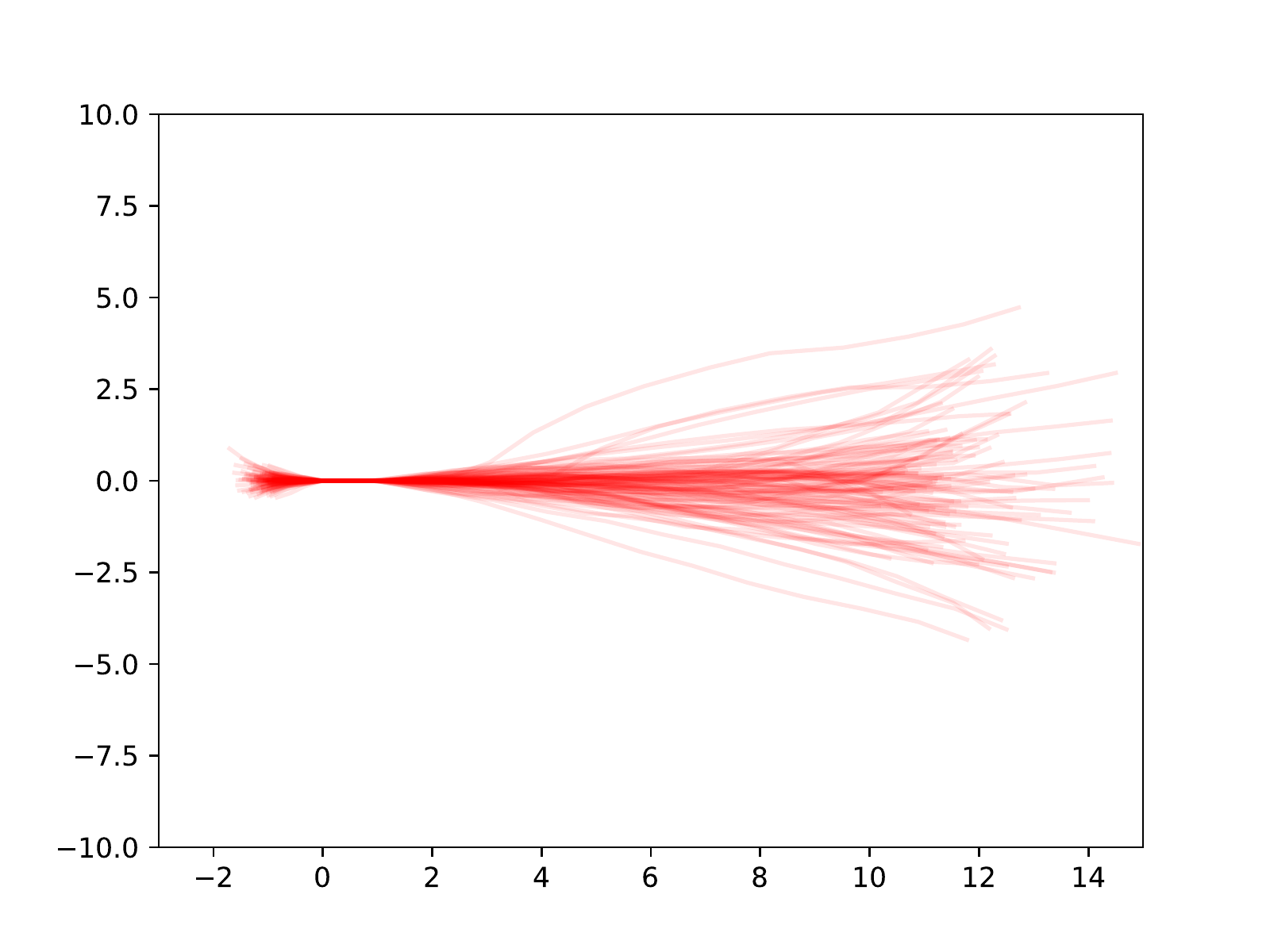}
\caption{}
\end{subfigure}
\begin{subfigure}{0.3\linewidth}
\centering
\includegraphics[width=1.0\linewidth]{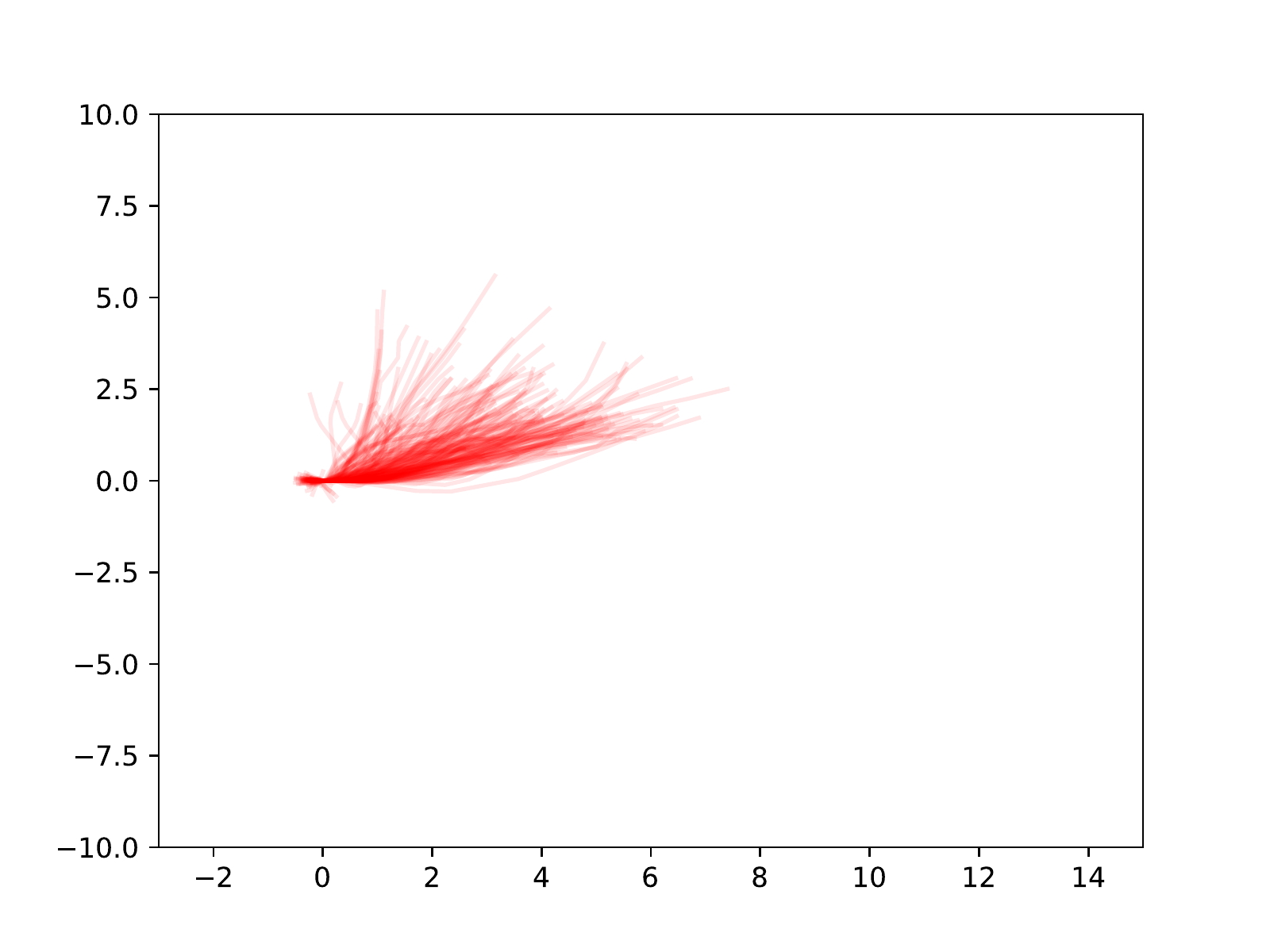}
\caption{}
\end{subfigure}
\begin{subfigure}{0.3\linewidth}
\centering
\includegraphics[width=1.0\linewidth]{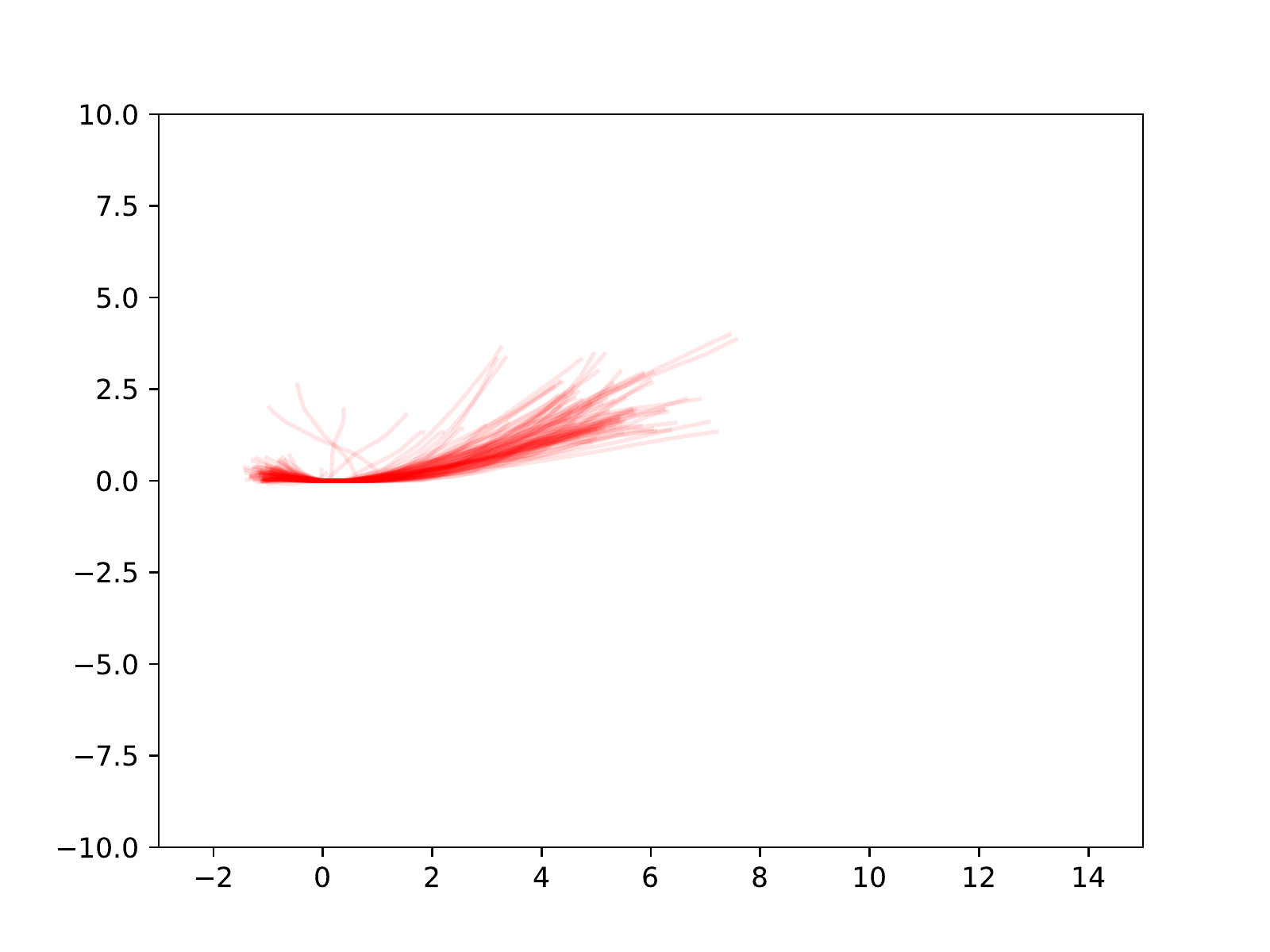}
\caption{}
\end{subfigure}
\begin{subfigure}{0.3\linewidth}
\centering
\includegraphics[width=1.0\linewidth]{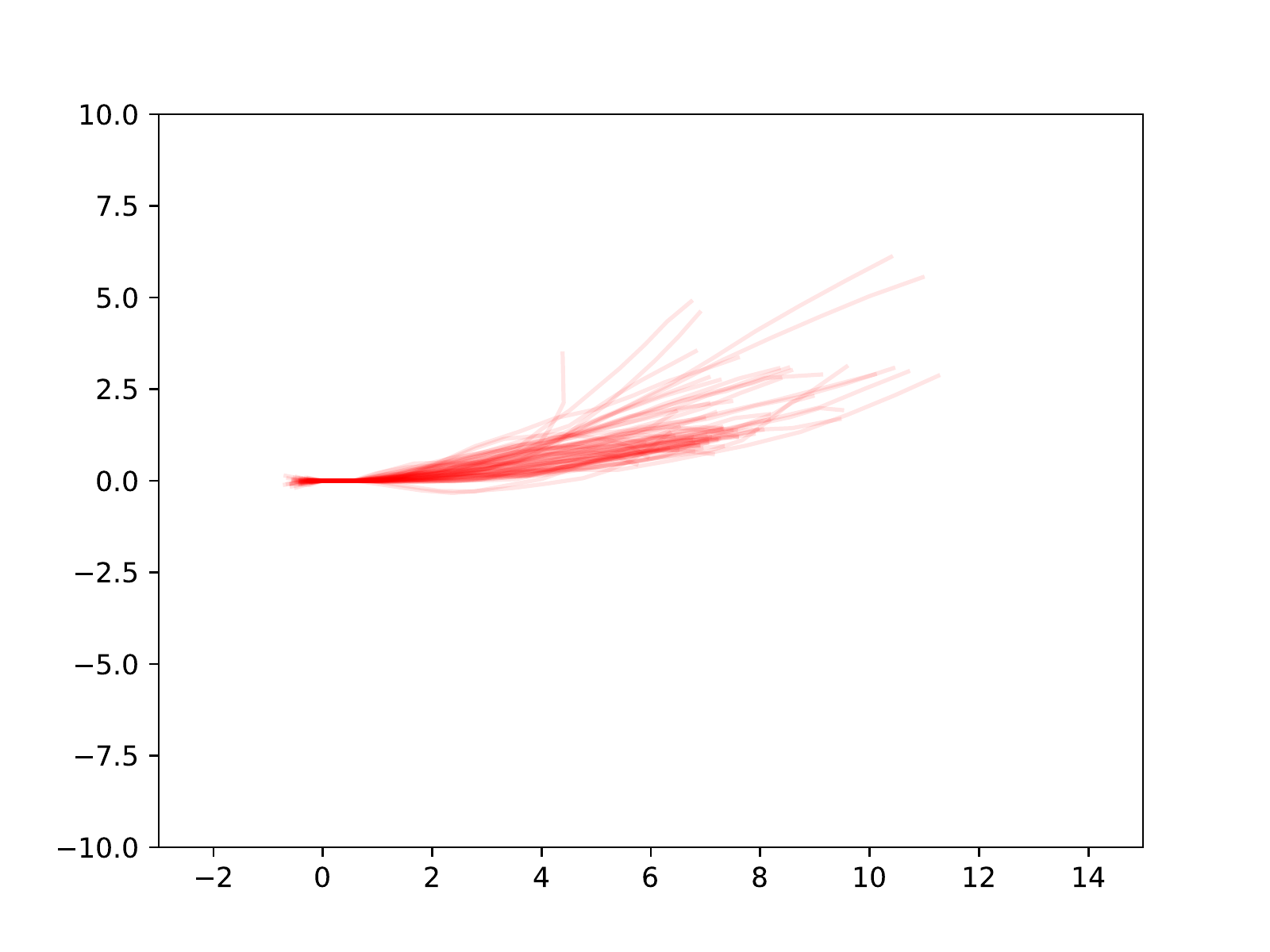}
\caption{}
\end{subfigure}
\begin{subfigure}{0.3\linewidth}
\centering
\includegraphics[width=1.0\linewidth]{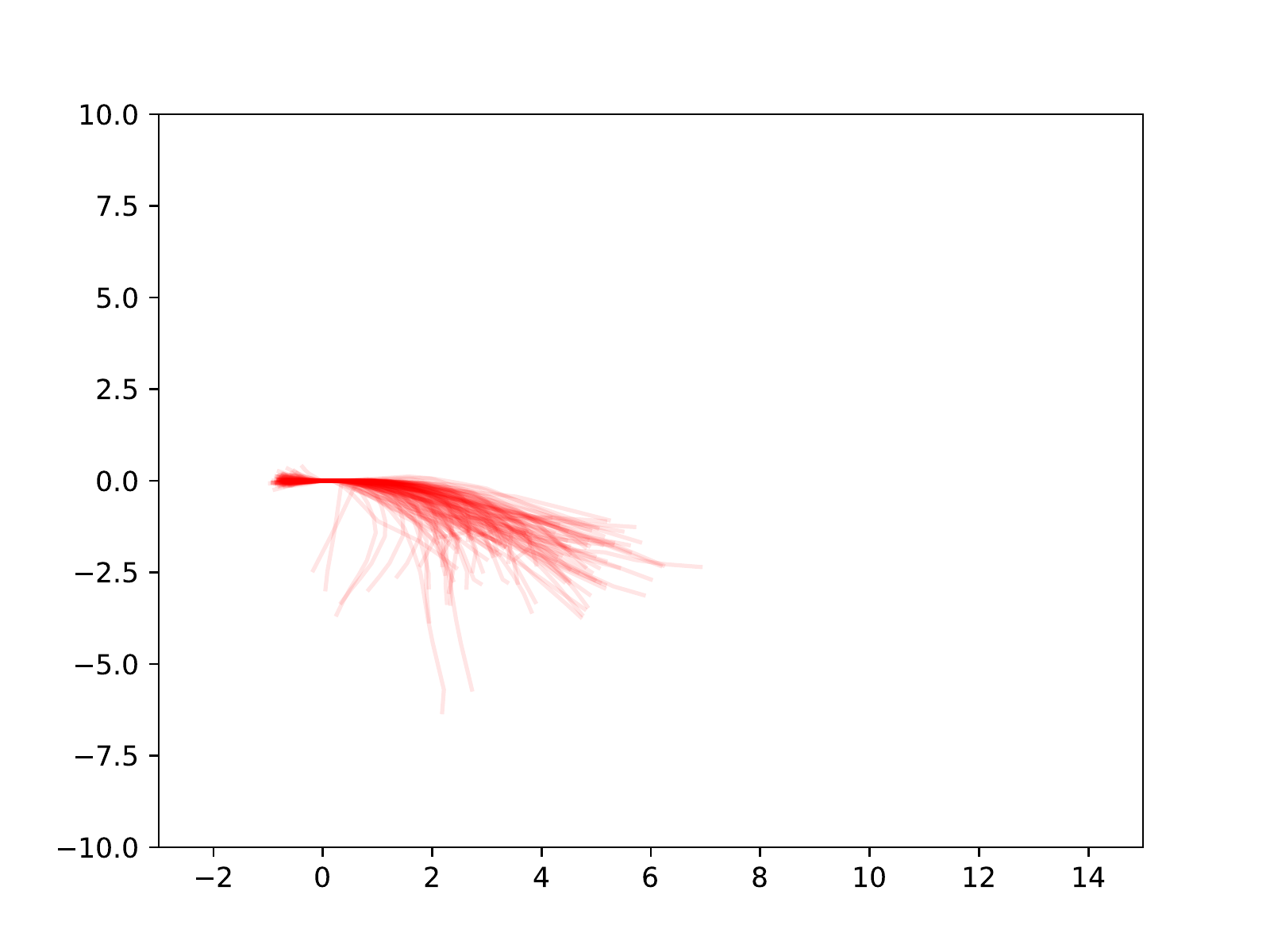}
\caption{}
\end{subfigure}
\begin{subfigure}{0.3\linewidth}
\centering
\includegraphics[width=1.0\linewidth]{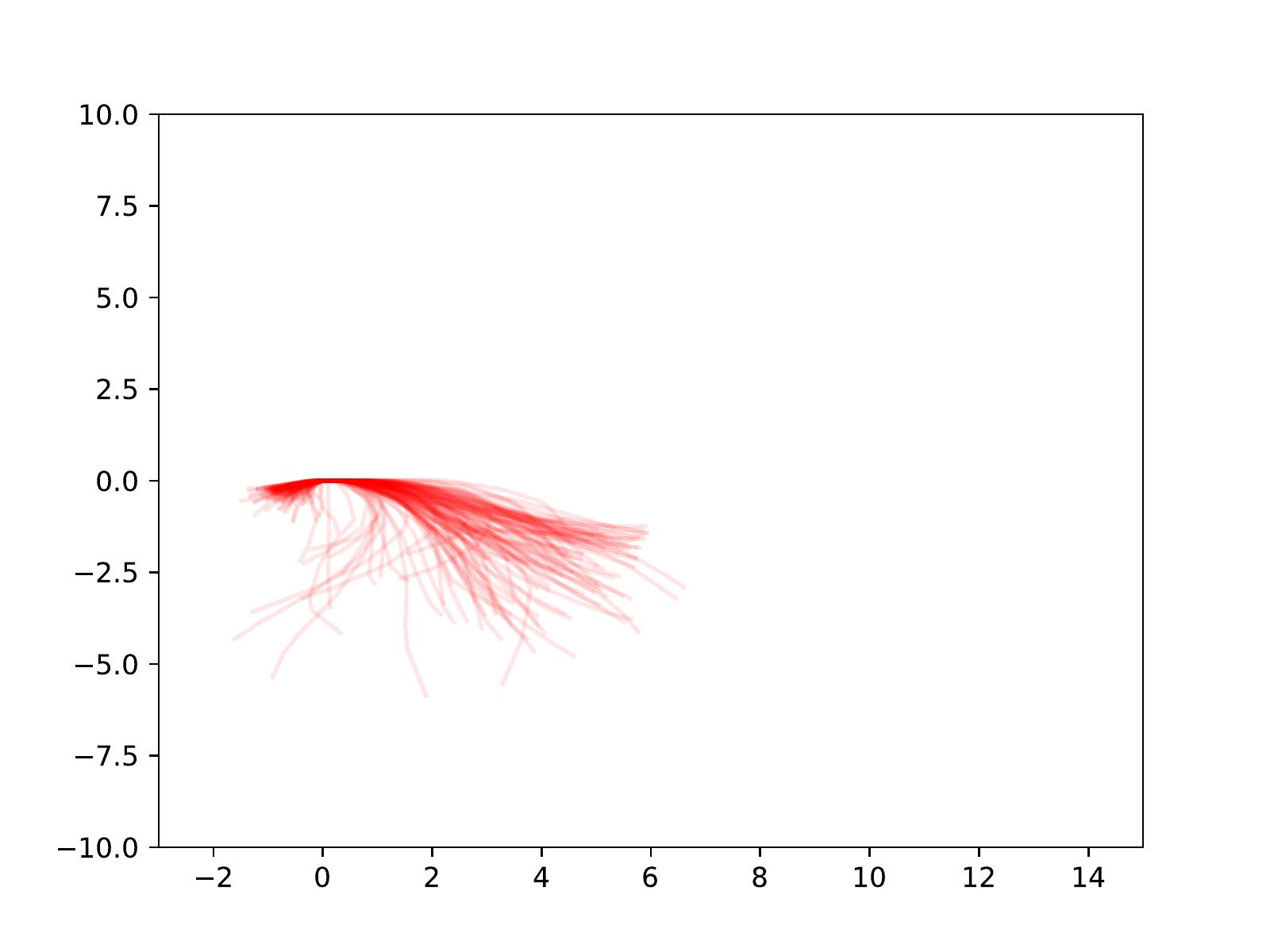}
\caption{}
\end{subfigure}
\begin{subfigure}{0.3\linewidth}
\centering
\includegraphics[width=1.0\linewidth]{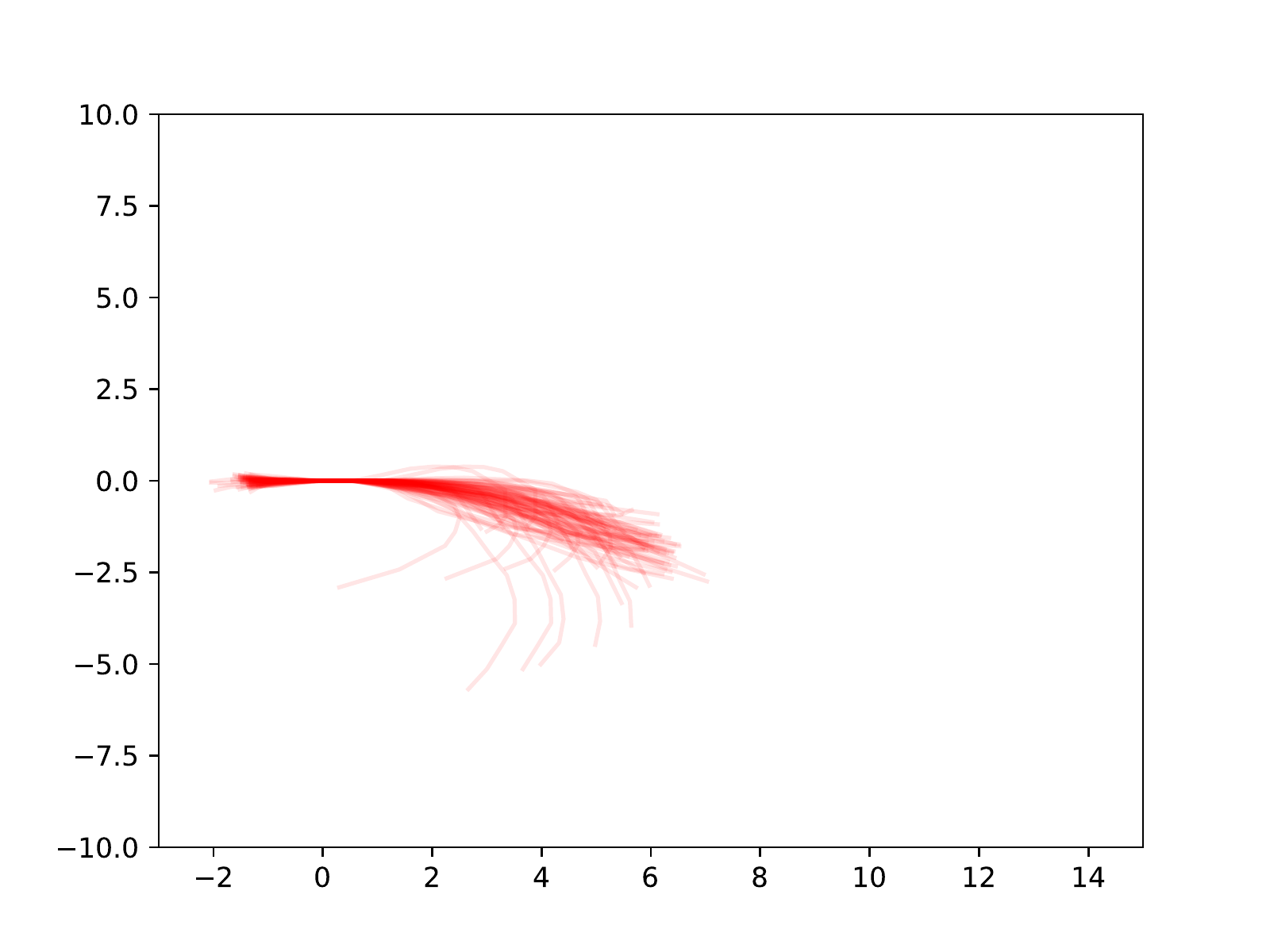}
\caption{}
\end{subfigure}
\caption{Trajectories in different tail prototype clusters.}
\label{fig:tail}
\end{figure}

In \cref{fig:head} we demonstrate different head trajectory prototype clusters on ETH-UCY. And in \cref{fig:tail} we demonstrate different tail trajectory prototype clusters. We can see that the tail trajectory prototypes are with various motion patterns and are hard to predict. 

\begin{figure}[]
\small
    \centering
    \includegraphics[width=1.0\linewidth]{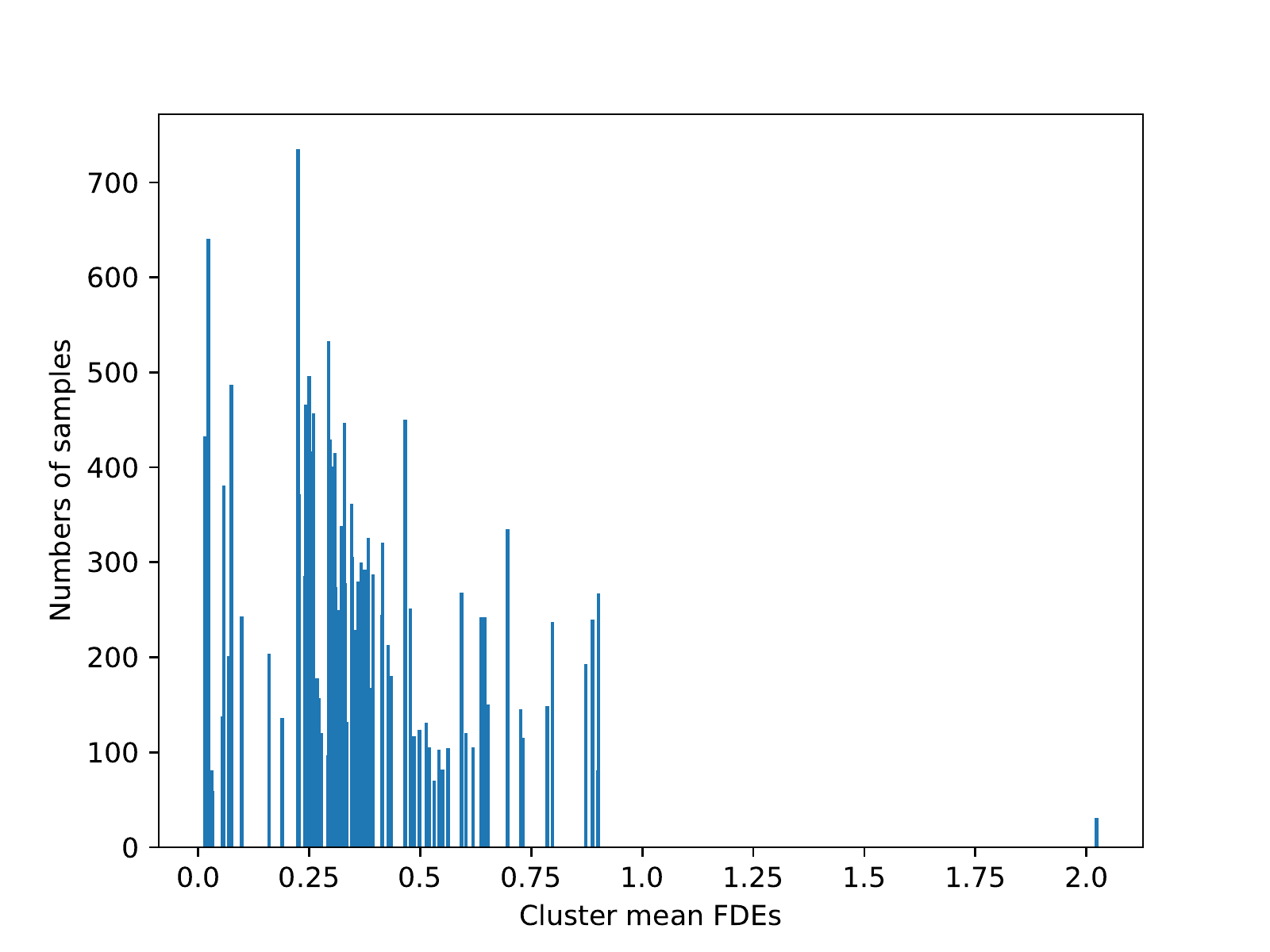}
    \caption{Distribution of the mean FDEs (from Traj++ EWTA) of different clusters. When two clusters have very similar FDEs, their bars will overlap.}
    \label{fig:cluster_error}
\end{figure}

Figure \ref{fig:cluster_error} shows the distribution of mean FDEs of the clusters for univ dataset. We can see that the samples are grouped into clusters by their hardness, generating some easy clusters with mean FDEs like $0.1$ and some hard clusters with mean FDEs like $0.9$ and $2.0$.

\section{Preliminary Implement Results On Another Baseline. }

In \cref{tab:stgcnn} we show a preliminary implementation of our method on another backbone: Social-STGCNN \cite{mohamed2020social}. The results in \cref{tab:stgcnn} are calculated by our implementation based on the public code of \cite{mohamed2020social}, and the results are calculated by taking an average of five individual runs considering the sampling randomness. We can see from \cref{tab:stgcnn} that our method can significantly improve the performances on all tailed samples, while the averaged performances are only very slightly affected. 
Those results show the generalization ability of our method. Doing contrastive learning will probably break the neighborhood correlations slightly, and we will further work on this problem in the future.  

\begin{table*}[]
\small
\centering
\begin{tabular}{lllllll}
\toprule
                   & Top 1\%            & Top 2\%            & Top 3\%            & Top 4\%            & Top 5\%            & All                \\
\midrule
Social-STGCNN \cite{mohamed2020social}     & 1.90/3.97          & 1.61/3.43          & 1.45/3.14          & 1.33/2.89          & 1.25/2.72          & \textbf{0.45/0.76} \\
Social-STGCNN+FEND & \textbf{1.73/3.69} & \textbf{1.49/3.23} & \textbf{1.35/2.96} & \textbf{1.27/2.76} & \textbf{1.20/2.59} & 0.46/0.78 \\
\bottomrule
\end{tabular}
\caption{Preliminary implement results of our FEND module on another baseline: Social-STGCNN. Results are in format of (minADE/minFDE) in meters. Bold numbers are the best results of each column. }
\label{tab:stgcnn}
\end{table*}

\section{More Visualization Results}

\begin{figure}[]
\centering
\begin{subfigure}{0.7\linewidth}
\centering
\includegraphics[width=1.0\linewidth]{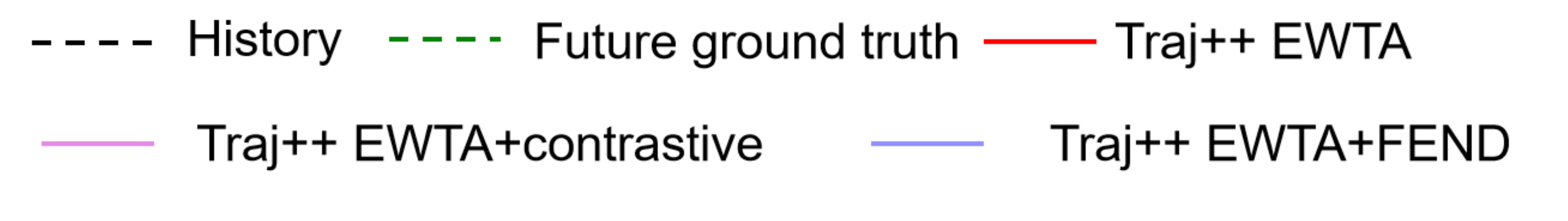}
\end{subfigure}
\begin{subfigure}{0.30\linewidth}
\centering
\fbox{\includegraphics[width=1.0\linewidth]{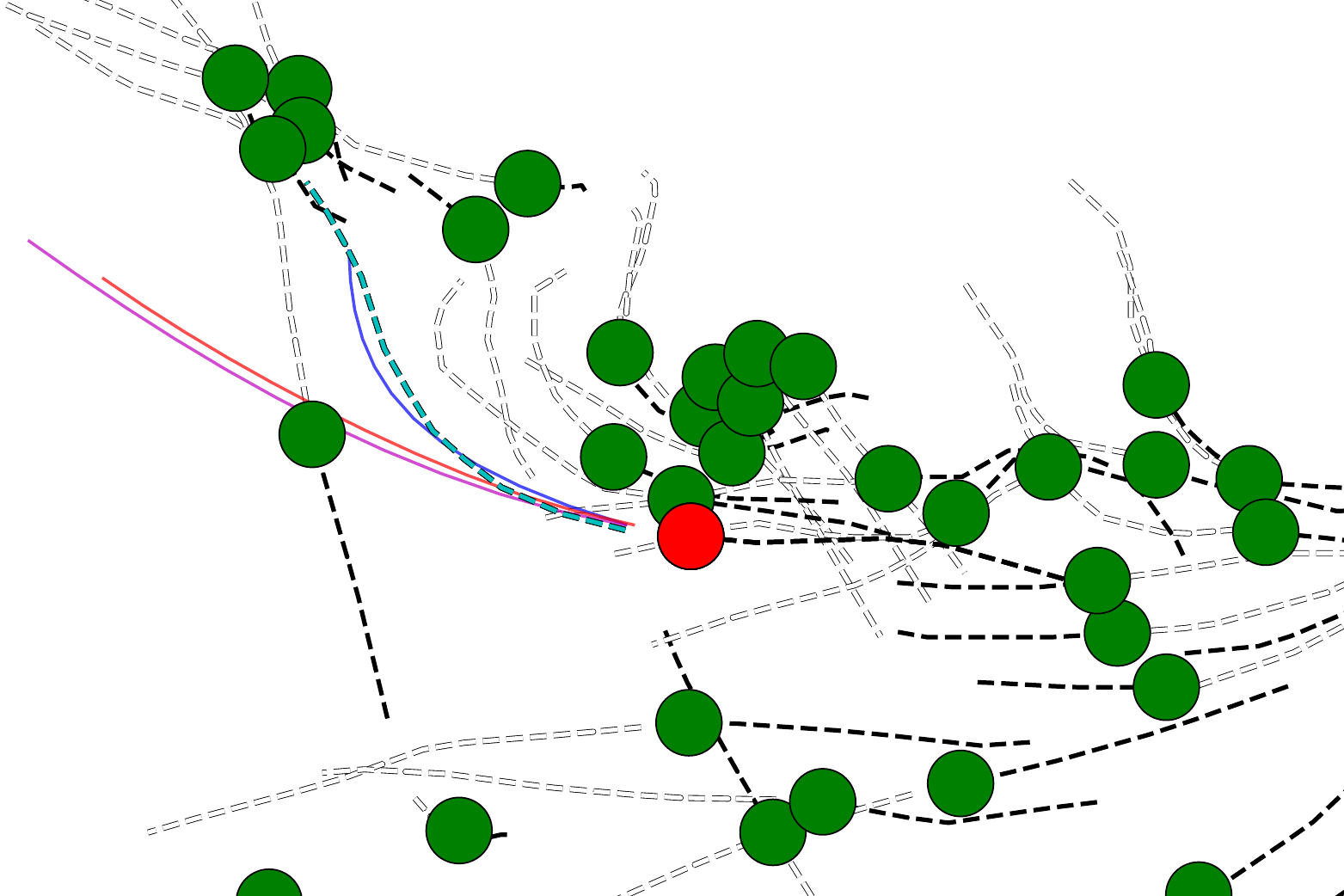}}
\caption{}
\end{subfigure}
\begin{subfigure}{0.3\linewidth}
\centering
\fbox{\includegraphics[width=1.0\linewidth]{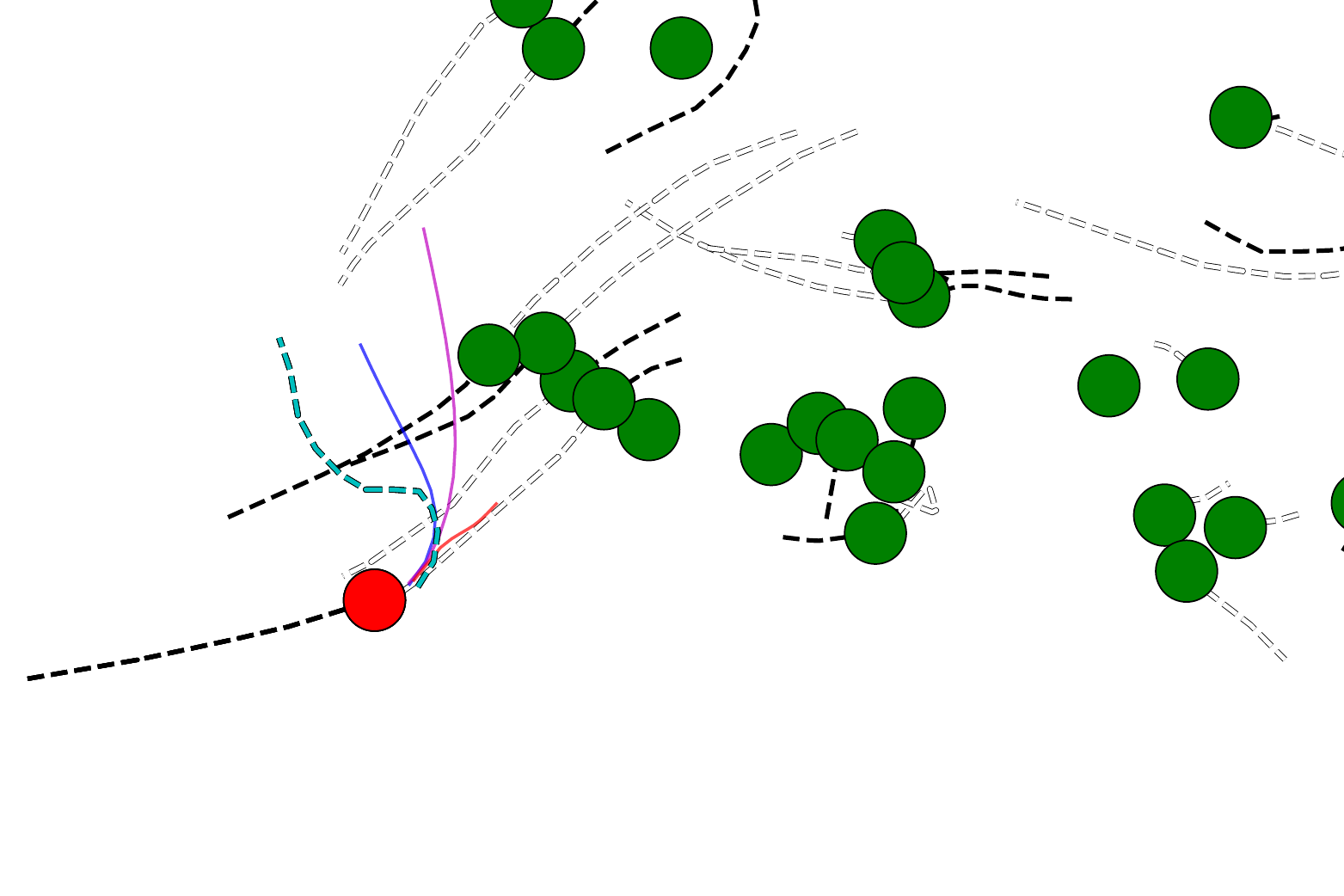}}
\caption{}
\end{subfigure}
\begin{subfigure}{0.3\linewidth}
\centering
\fbox{\includegraphics[width=1.0\linewidth]{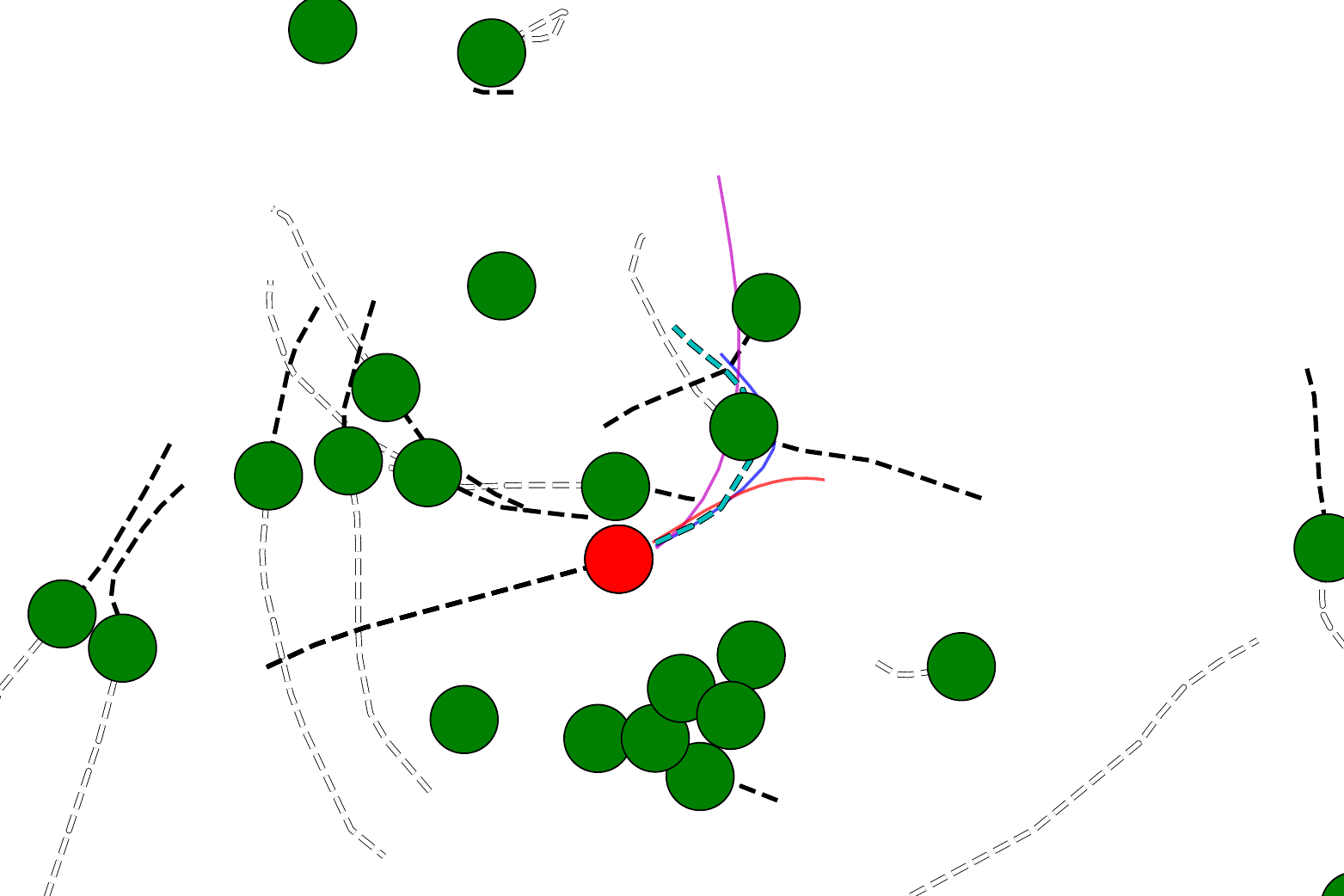}}
\caption{}
\end{subfigure}
\begin{subfigure}{0.3\linewidth}
\centering
\fbox{\includegraphics[width=1.0\linewidth]{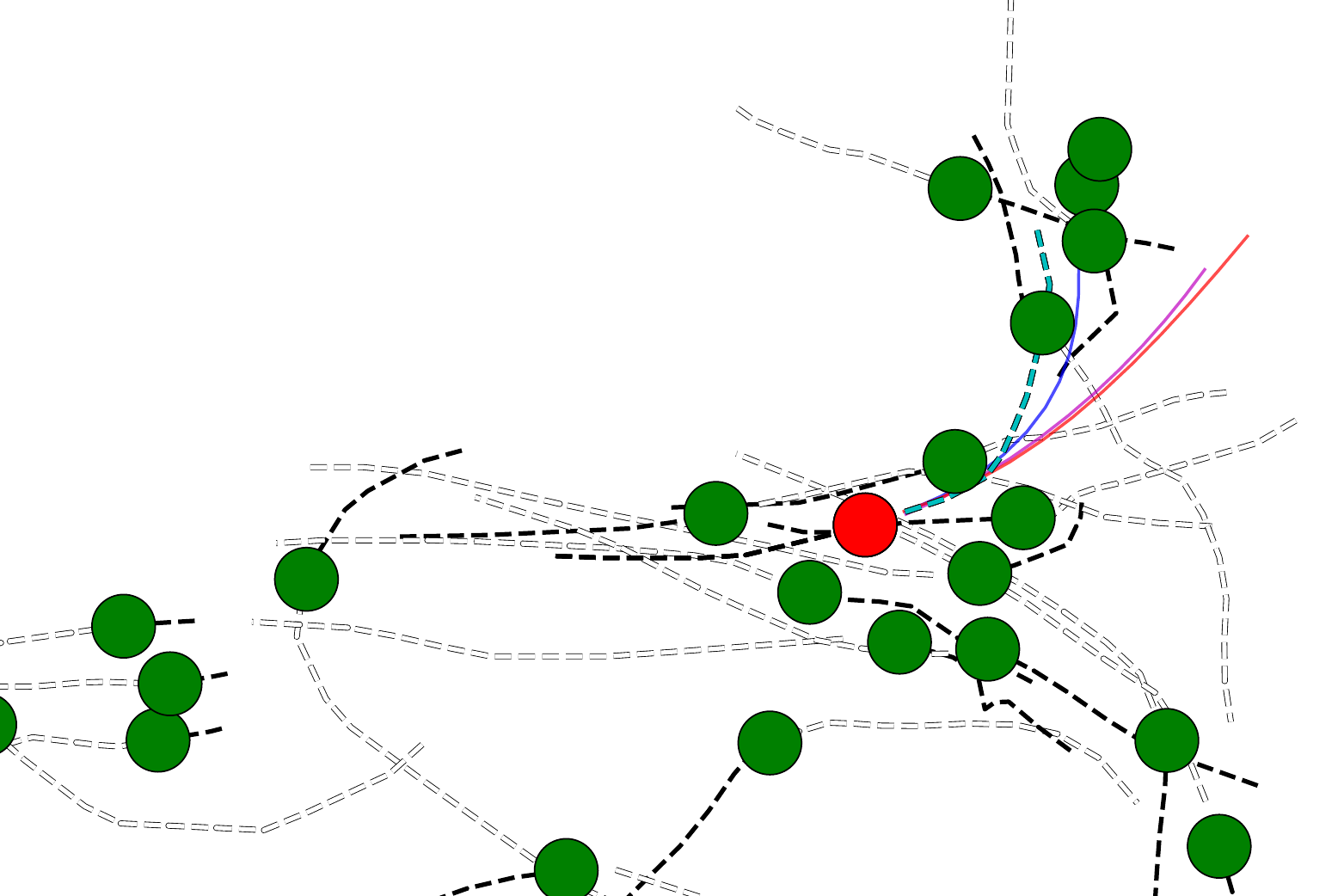}}
\caption{}
\end{subfigure}
\begin{subfigure}{0.3\linewidth}
\centering
\fbox{\includegraphics[width=1.0\linewidth]{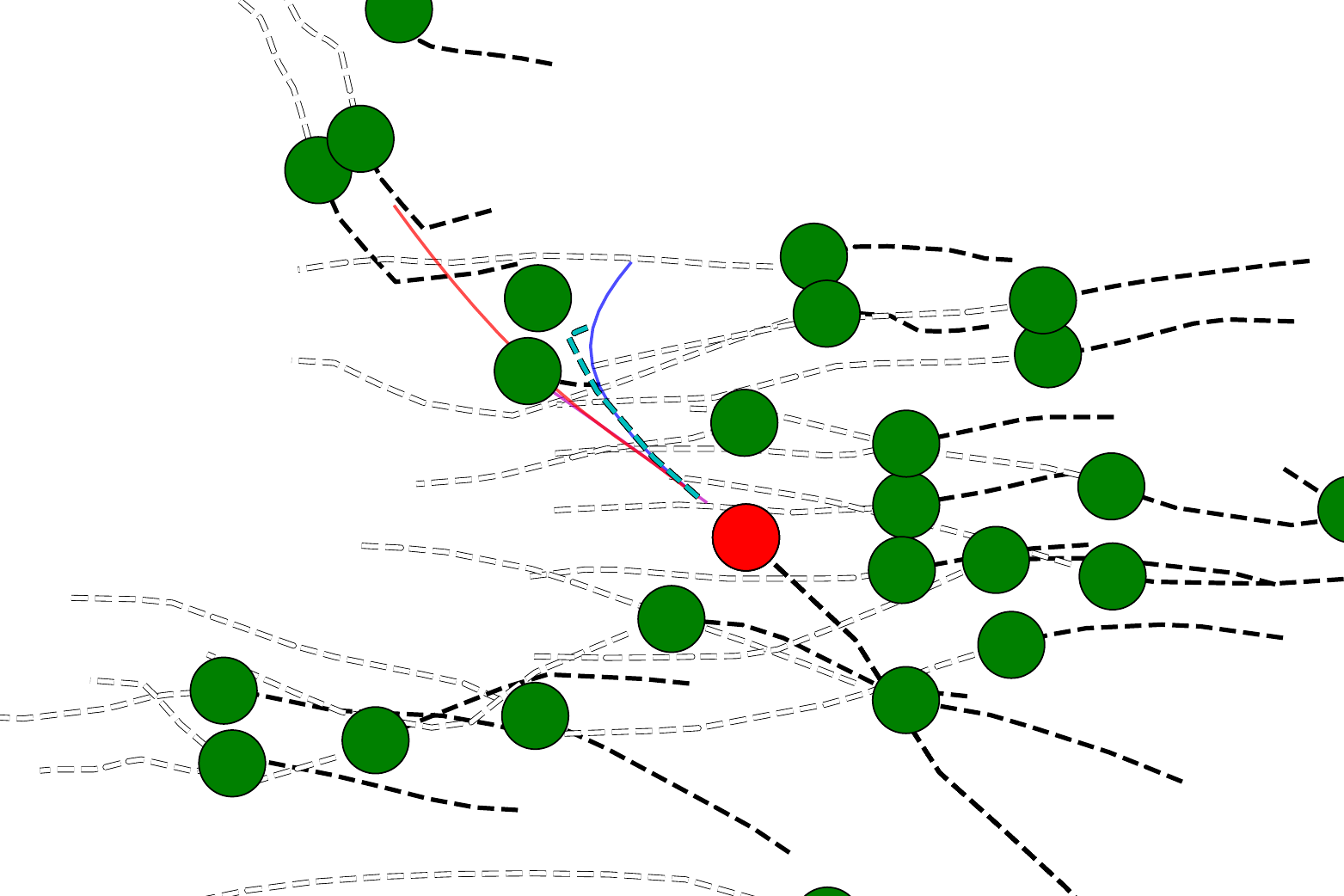}}
\caption{}
\end{subfigure}
\begin{subfigure}{0.3\linewidth}
\centering
\fbox{\includegraphics[width=1.0\linewidth]{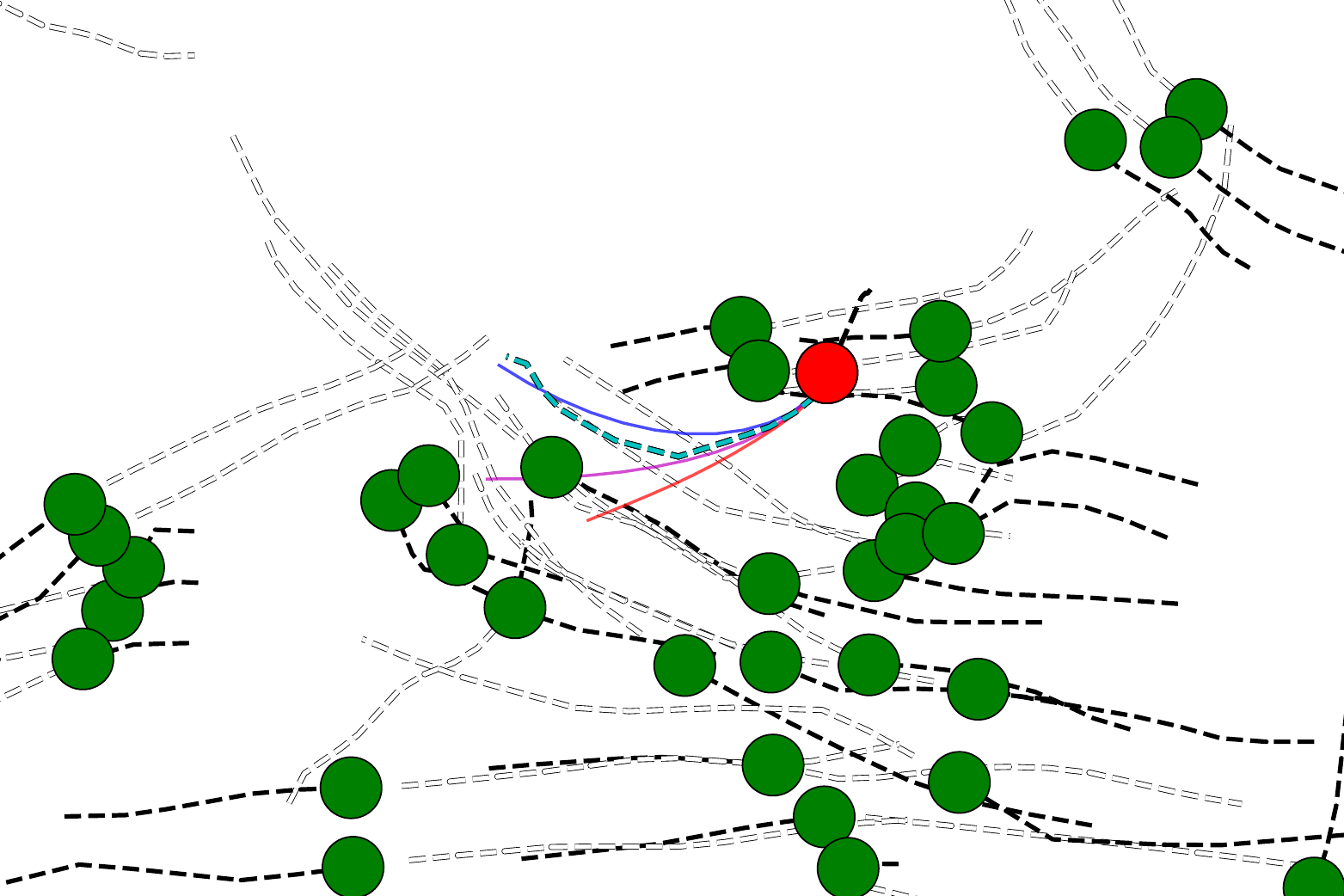}}
\caption{}
\end{subfigure}

\caption{Qualitative visualizations of our model versus Traj++ EWTA and Traj++ EWTA + contrastive .}
\label{fig:vis}
\end{figure}

In \cref{fig:vis} we demonstrate more qualitative comparisons of our method in ETH-UCY dataset with Traj++ EWTA \cite{makansi2021exposing} and Traj++ EWTA + contrastive \cite{makansi2021exposing}. We can see that in all those challenging scenarios our method outperforms those two methods.

\begin{table}[]
\small
\centering
\begin{tabular}{llllllll}
\toprule
$\theta$& Top 1\%     & Top 3\%     & Top 5\%   & All         \\ 
\midrule
0   & \textbf{0.84/2.12}  & 0.62/1.48 & 0.53/1.21 & 0.17/0.32 \\ 
0.2 & 0.84/2.13  & \textbf{0.61/1.46} & \textbf{0.52/1.19} & \textbf{0.17/0.32} \\ 
0.5 & 0.91/2.32  & 0.64/1.58 & 0.54/1.27 & 0.17/0.32 \\ 
\bottomrule
\end{tabular}
\caption{Study on the parameter sensitivity of the threshold $\theta$ .Results are in the format of (minADE/minFDE) in meters.}
\label{tab:parameter study}
\end{table}

\section{More Parameter Sensitivity Study}

Table \ref{tab:parameter study} shows the parameter study of the head sample filter threshold $\theta$ to apply PCL auxiliary loss. We can see that adding $\theta=0.2$ outperforms the model without a threshold, but filtering out too much samples will harm the performance of the tail samples, as $\theta=0.5$ shows.

\section{Performances On Kalman Filter Scores}
Our method can also perform decently on another hardness score: Kalman Filter prediction errors. 
Table \ref{tab:kalman} contains the results for the Kalman scores versus \cite{makansi2021exposing}, which is the state-of-the-art long-tail trajectory prediction method. T++E means Traj++ EWTA. Note that the Top $1\%$ hard samples selected by Kalman prediction scores have a lower mean FDE than the Top $2-5\%$ samples, indicating that the Kalman errors can not reveal the sample hardness with respect to the neural network predictor.

\begin{table*}[h]
\small
\centering
\begin{tabular}{lllllll}
\toprule

                  & Top 1\%       & Top 2\%      & Top 3\%  & Top 4\%  & Top 5\%   & All       \\ 
\midrule
T++E contrastive \cite{makansi2021exposing} & 0.38/\textbf{0.71}  & 0.48/1.03 & 0.46/1.03 & 0.45/1.00   & 0.42/0.90  & 0.17/0.32 \\ 
T++E FEND         & \textbf{0.38}/0.74  & \textbf{0.43}/\textbf{0.92} & \textbf{0.40}/\textbf{0.85} & \textbf{0.39}/\textbf{0.82} & \textbf{0.37}/\textbf{0.76} & \textbf{0.17}/\textbf{0.32} \\ 
\bottomrule
\end{tabular}
\caption{Prediction errors on hard tail samples selected by Kalman prediction errors on ETH-UCY. Results are in the format of (minADE/minFDE) in
meters. Bold numbers mean the best results of each column.}
\label{tab:kalman}
\end{table*}

\section{More Ablation}

\begin{table*}[]
\small
\centering
\begin{tabular}{lllllll}
\toprule
                           & Top 1\%            & Top 2\%            & Top 3\%            & Top 4\%            & Top 5\%            & All                \\
\midrule
Traj++ EWTA\cite{makansi2021exposing}                      & 0.98/2.54 & 0.79/2.07 & 0.71/1.81 & 0.65/1.63 & 0.60/1.50  & \textbf{0.17/0.32} \\ 
Traj++ EWTA + hypernetwork & 0.97/2.46          & 0.78/2.00          & 0.69/1.72          & 0.62/1.54          & 0.57/1.40          & 0.17/0.33          \\
Traj++ EWTA + hypernetwork +PCL    & 0.90/2.28          & 0.72/1.87          & 0.65/1.61          & 0.58/1.43          & 0.54/1.30                & \textbf{0.17/0.32}          \\ 
Traj++ EWTA + FEND         & \textbf{0.84/2.13} & \textbf{0.68/1.68} & \textbf{0.61/1.46} & \textbf{0.56/1.30} & \textbf{0.52/1.19} & \textbf{0.17/0.32} \\
\bottomrule
\end{tabular}
\caption{Ablation studies of using the hypernetwork. Results are in format of (minADE/minFDE) in meters. Bold numbers are the best results of each column. }
\label{tab:hyper}
\end{table*}

In \cref{tab:hyper} we show more ablation results of our method with regard to the hypernetwork. We can see that using both hypernetwork and PCL can help with improving performances on tail samples.

\section{Baseline Implement Details}
We do LDAM \cite{cao2019learning} experiments using the suggested setting in \cite{makansi2021exposing}. The dataset
is divided into 13 classes according to Kalman errors. Then
an MLP classification head is added after the trajectory embeddings to predict the class divisions as an auxiliary task. The main hyperparameters are $s$ in LDAM loss and $w$ for task loss reweighing. We tried different values of the main
hyperparameters, and finally use the $s = 1$, $w = 1$ setting, the same as the suggestion in \cite{makansi2021exposing}.

\begin{table}[]
\small
\centering
\begin{tabular}{lllll}
\toprule
         & Top 1\% & Top 2\% & Top 3\% & All   \\ 
\midrule
T++ \cite{salzmann2020trajectron++}     & 8.23 & 6.37 & 5.52 & -1.12 \\ 
T++ FEND & \textbf{7.97} & \textbf{6.23} & \textbf{5.31} & \textbf{-1.12} \\ 
\bottomrule
\end{tabular}
\caption{Full FDE NLL results on Trajectron++ on ETH-UCY. Bold numbers are the best results of each column.}
\label{tab:FDE NLL}
\end{table}

\section{Performances On Another Metric}
Except the minADE/FDE, our method is a plug-in module and can be evaluated
with any better baselines and metrics. We plug our module
into a probabilistic method : Trajectron++ (T++) \cite{salzmann2020trajectron++}, and evaluate using full KDE NLL. The result is in Table \ref{tab:FDE NLL}.

\begin{table}[]
\footnotesize
\centering
\begin{tabular}{lllll}
\toprule
                         & Top 1\%            & Top 2\%            & Top 3\%            & All                \\
\midrule
T  \cite{makansi2021exposing}                  & 1.85/4.63          & 1.44/3.69          & 1.23/3.15          & 0.19/0.34          \\
T+contrastive        & 1.45/3.27          & 1.11/2.54          & 0.95/2.17          & 0.19/0.32          \\
T w/o RS+FEND & \textbf{1.24/2.48} & \textbf{0.94/1.88} & \textbf{0.81/1.63} & \textbf{0.18/0.27} \\
\bottomrule
\end{tabular}
\caption{Results of 6 timesteps prediction on Nuscenes. Results are in format of (minADE/minFDE). Bold numbers are the best
results of each column. }
\label{tab:test 6}
\end{table}

\section{Performances On Different Prediction Time Periods on Nuscenes}
We found out that the Trajectron++ EWTA model is
trained with 6 as the prediction horizon but tested with 8
as the prediction horizon. We also provide the results for
testing with 6 future timesteps in \cref{tab:test 6}. T means Trajectron++ EWTA.

\end{document}